%% file: main.tex
\pdfoutput=1

\documentclass[11pt]{article}
\usepackage[T1]{fontenc}
\usepackage[utf8]{inputenc}
\usepackage[dvipsnames]{xcolor}

\usepackage{acl}

\usepackage{times}
\usepackage{latexsym}
\usepackage{booktabs}
\usepackage[inline]{enumitem}
\usepackage{multirow}
\usepackage{makecell}
\usepackage{graphicx}
\usepackage{amsmath}
\usepackage{standalone}
\usepackage{tikz}
\usepackage{caption}
\usepackage{subcaption}
\usepackage{pgfplots}
\usepackage{tcolorbox}
\usepackage{microtype}

\pgfplotsset{compat=newest}
\usepgfplotslibrary{statistics}
\usetikzlibrary{arrows, matrix, positioning, math, external, patterns, backgrounds, fit, calc,  backgrounds, shapes.callouts}

\tikzset{
  text_box/.style = {rectangle callout, rounded corners=5pt, draw=none, text width=0.5\textwidth},
}

\newtcolorbox[auto counter,number format=\arabic]{exbox}[2][]{%
    colback=red!5!white,
    boxrule=0.3pt,
    colframe=black,
    fonttitle=\bfseries,
    boxsep=1mm,
    arc=1mm,
    halign=justify,
    width=\columnwidth,
    title=Example~\thetcbcounter: #2,#1}

\definecolor{bblue}{HTML}{5cb1eb}
\definecolor{rred}{HTML}{be1908}
\definecolor{ggreen}{HTML}{aaad00}
\definecolor{dblue}{HTML}{001158}
\definecolor{ibm1}{HTML}{0077BB}
\definecolor{ibm2}{HTML}{33BBEE}
\definecolor{ibm3}{HTML}{EE7733}
\definecolor{ibm4}{HTML}{EE3377}
\definecolor{ibm5}{HTML}{CC3311}
\definecolor{seq1}{HTML}{1965B0}
\definecolor{seq2}{HTML}{7BAFDE}
\definecolor{seq3}{HTML}{4EB265}
\definecolor{seq4}{HTML}{F7F056}
\definecolor{seq5}{HTML}{DC050C}

\newcolumntype{P}[1]{>{\centering\arraybackslash}p{#1}}

\input{commenting_macros.tex}

\title{Do Differences in Values Influence Disagreements in Online Discussions?}

\author{Michiel van der Meer\\
  LIACS \\ Leiden University \\
  \texttt{\small m.t.van.der.meer@liacs.leidenuniv.nl} \\\And
  Piek Vossen \\
  CLTL \\ Vrije Universiteit Amsterdam \\
  \texttt{\small p.t.j.m.vossen@vu.nl} \\\AND
  Catholijn M. Jonker \\
  Interactive Intelligence \\ TU Delft \\
  \texttt{\small c.m.jonker@tudelft.nl} \\\And
  Pradeep K. Murukannaiah \\
  Interactive Intelligence \\ TU Delft \\
  \texttt{\small p.k.murukannaiah@tudelft.nl}}

\begin{document}
\maketitle
\begin{abstract}
Disagreements are common in online discussions. Disagreement may foster collaboration and improve the quality of a discussion under some conditions. Although there exist methods for recognizing disagreement, a deeper understanding of factors that influence disagreement is lacking in the literature.
We investigate a hypothesis that differences in \emph{personal values} are indicative of disagreement in online discussions. We show how state-of-the-art models can be used for estimating values in online discussions and how the estimated values can be aggregated into value profiles. We evaluate the estimated value profiles based on human-annotated agreement labels.
We find that the dissimilarity of value profiles correlates with disagreement in specific cases. We also find that including value information in agreement prediction improves performance.
\end{abstract}

\section{Introduction}
A large number of users participate in online deliberations on societal issues such as climate change \citep{beel2022linguistic} and vaccination hesitancy \citep{weinzierl2022hesitancy}. Disagreement is an important aspect of a deliberation \citep{polletta2018forms} since it can
\begin{enumerate*}[label=(\arabic*)]
    \item drive novel ideas,
    \item incentivize evaluation of the proposed ideas,
    \item avoid echo chambers, and
    \item cancel out individual biases \citep{klein2012enabling}.
\end{enumerate*}
Discussions with disagreement help users understand the opposing viewpoints \citep{lin2022learning,saveski2022perspective}. Further, discussions having adequate disagreement have been associated with a higher quality deliberation \citep{esterling2015much}.

Ensuring that participants express a sufficient level of disagreement in a discussion is hard. We do not know the nature of disagreement in online platforms \citep{stromer2020context}.
Further, questions arise on how to control for disagreement to enhance reciprocity \citep{esau2022creates}, and how too much exposure to opposing views drives polarization \citep{bail2018exposure}.
Analysis methods for online discussions currently cannot accurately represent such diverse perspectives \citep{cabitza2023toward, vandermeer2022hyena}, and measuring deliberative quality is an open challenge \citep{vecchi2021towards,shortall2022reason}.

We want to ensure that a discussion incorporates many perspectives and that those are actively communicated. For this reason, we turn to \emph{value conflicts}, a potential root cause for disagreement. We consider the hypothesis that when users with conflicting values engage in a discussion, diverging views come up.
Perspective and value clashes are at the heart of disagreement \citep{stromer2009agreement}. In collaborative teams, value conflicts are linked to disagreement \citep{jehn1994enhancing}. Specifically, values are said to be an effective way to make conflict explicit among participants in a discussion \citep{beck2019managing}.

\begin{figure*}[!htb]
    \centering
    \includegraphics[width=\textwidth]{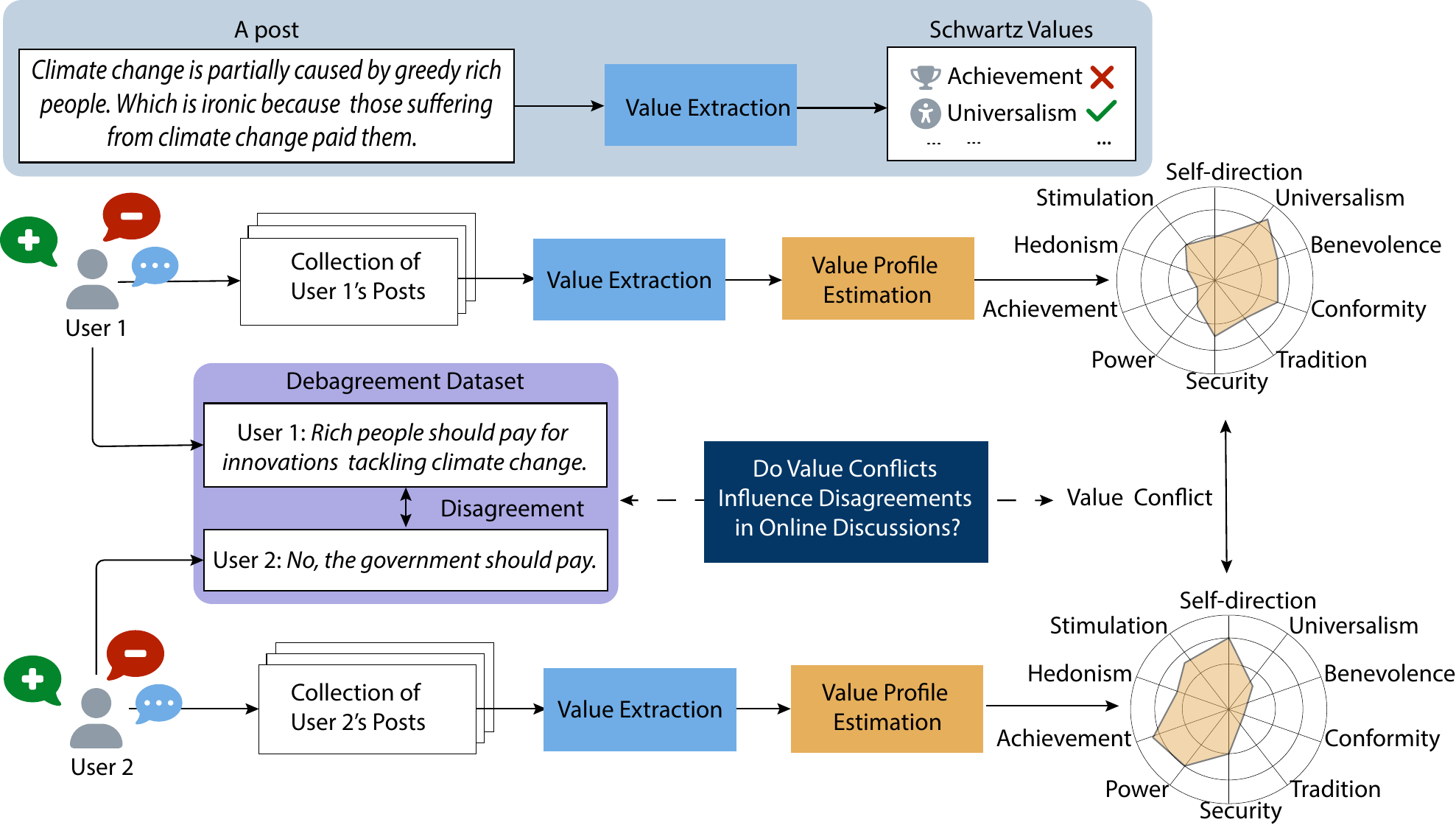}
    \caption{Setup of measuring value conflicts by means of Value Profile Estimation (VPE).}
    \label{fig:vpe-setup}
\end{figure*}

To evaluate our hypothesis, we construct value profiles based on user comments on Reddit, a social media platform. A value profile captures the relative importance a user ascribes to values. We employ ten values, e.g., stimulation, universalism, and security, from the well-known Schwartz theory of basic values \cite{schwartz2012overview}. Then, we compare the similarities among profiles to the disagreement among users on different topics. This allows us to investigate the association between value conflict (low similarity) and disagreement. Figure~\ref{fig:vpe-setup} shows an overview of our approach.

We gather 11.4M comments from 19K users on Reddit to construct value profiles. We perform up to 200 tests with different settings to investigate our hypothesis. We further experiment with replacing estimated value profiles with self-reported ones. To do so, we collect 572 judgments from 26 annotators in combination with self-reported value profiles.

Selecting conversation partners based on their profile to manage value conflicts and influence the level of disagreement in a discussion could be a tool for moderators to balance conversations.
To provide support for moderators, we investigate the impact of adding profile information to the agreement analysis task \citep{pougue2021debagreement}. Since the contextual implications of values are usually unknown, connecting user concerns to values \citep{alshomary2022moral} opens up human-machine collaboration opportunities for a more constructive conversation \citep{akata2020research, hadfi2022augmented, liscio2022values}.

\paragraph{Contributions}
\noindent\begin{enumerate*}[label=(\arabic*)]
    \item We experiment with methods for value estimation from text to obtain value profiles from an online discourse (Reddit comments).
    \item We investigate how value conflicts affect disagreement in discussions by showing that low-profile similarity can co-occur with disagreement under specific conditions for estimated and self-reported value profiles.
    \item We make first steps in using the value-laden background information for predicting user disagreement and comparing it to other user-specific contextual information.
\end{enumerate*}

\section{Related Work}
Although there is existing work on analyzing agreement in online discussions, very few works focus on examining the reasons for disagreement. We review the existing work on agreement analysis, introduce two popular value theories, and outline previous research on value estimation.

\subsection{(Dis)-agreement and discussion analysis}
Detecting whether people agree or disagree with given statements is commonly framed as stance classification \citep[e.g.,][]{aldayel2021stance}.
Recently, more effort has been put into exploring various aspects of the task \citep{hardalov2021cross,allaway2020zero,liu2021enhancing}.
However, little work is done in adjusting the task to detect stances among users within online discussions. To this end, \textbf{agreement analysis} focuses on detecting (dis-)agreement in data that
\begin{enumerate*}[label=(\arabic*)]
    \item represents realistic online discussions,
    \item provides contextual information (post authors, timestamps, etc.),
    \item contains diverse writing styles,
    \item touches on multiple topics \citep{pougue2021debagreement}.
\end{enumerate*}

Existing work on agreement analysis is aimed at
\begin{enumerate*}[label=(\arabic*)]
    \item identifying language that indicates disagreement \citep[e.g.,][]{niculae2016conversational, wojatzki2018agree, fischer2022typology},
    \item leveraging stylistic choices like sarcasm for detecting disagreement \citep{ghosh2021laughing},
    \item finding stance and target pairs, followed by the traditional stance classification \citep[e.g.,][]{chen2019seeing, de2021beg}, and
    \item mixing detailed opinion information using e.g., logic of evaluation \citep{draws2022comprehensive}.
\end{enumerate*}
Recently, adding social role context to textual comments was shown to have a positive impact on the agreement analysis task \citep{luo2023improving}, which indicates the usefulness of background information.
In this work, we focus on capturing the implicit motivations underlying opinions using \emph{personal values}, which have been known to drive individual opinions and actions across cultures \citep{schwartz2012overview}.

\subsection{Value models}
Values explain ideological beliefs underlying actions and opinions and may guide the design of applications \citep{friedman2013value}. Two leading value models have been used in NLP research: Moral Foundations \citep{graham2013moral} and the Schwartz Value model \citep{schwartz2012overview}. Each of these models includes a set of general values.

The Moral Foundation Theory (MFT) includes five foundations, each a vice--virtue dichotomy (e.g., \emph{harm}--\emph{care}). However, MFT does not stipulate any relationship among the foundations. In contrast, the Schwartz model includes ten basic values organized as a circumplex (right-hand side of Figure~\ref{fig:vpe-setup}), where similar values are placed close to each other. Further, Schwartz values can be grouped into four classes: \emph{openness to change}, \emph{conservation}, and \emph{self-transcendence}, \emph{self-enhancement}. Since the Schwartz model has more values and a structure among the values, it is better suited than MFT for comparing the value profiles of individuals. Thus, we employ Schwartz values in our work.

\subsection{Value estimation}
Most works based on representing an individual's value priorities (value profiles) use explicit preference elicitation, such as self-reporting and questionnaires \citep[e.g.,][]{boyd2015values}. However, a promising behavior-based approach focuses on analyzing textual motivations \citep{chen2014understanding}. To this end, dictionary-based approaches can be used for finding value mentions in texts \citep{ponizovskiy2020development, graham2009liberals}. Using such lexicons shows promising results in large-scale value estimation applications \citep{silva2021predicting}.

Recently, datasets annotated with personal values for training NLP methods have been released. In this paper, we use two recent datasets annotated with Schwartz values:
\begin{enumerate*}[label=(\arabic*)]
    \item ValueNet \citep{qiu2021valuenet} is a dataset containing textual scenarios related to moral decision-making that have been annotated with relevant Schwartz values.
    \item ValueArg \citep{kiesel2022identifying} contains user-submitted arguments that relate to specific Schwartz values.
\end{enumerate*}

There are some datasets on MFT values, e.g., \cite{trager2022moral, lourie2021scruples, hoover2020moral}. These datasets include value annotations for messages but do not include a link between the messages and users. Thus, estimating value profiles from such datasets is not possible.

Applications include dialogues about moral scenarios \citep{qiu2021valuenet}, review texts \citep{obie2021first}, and value-laden arguments \citep{kobbe2020exploring, alshomary2022moral}. However, both the annotation and extraction of values remain difficult, with specific questions relating to the granularity of the value labels \citep{kiesel2022identifying}, their transfer to new domains \citep{liscio2022cross}, and how classifiers understand morality in language \citep{liscio2023does}. Moreover, large variances exist between the frequency of values across domain \citep{kennedy2021moral}, and even the relevance of values differs depending on the domain \citep{bouman2018measuring, Liscio-2021-AAMAS-Axies}. However, users can still be represented inside each domain by examining relative frequencies inside value profiles, as stipulated by \citet{schwartz2012overview}.

\section{Method}
Figure~\ref{fig:vpe-setup} shows an overview of our method. We collect posts from users in online discussions. Using a trained value estimation model, we aggregate predictions over the collection to form a value profile. Then, to evaluate our hypothesis, we compare the value profiles for users known to be in disagreement based on an existing dataset. Our code\footnote{\url{https://github.com/m0re4u/value-disagreement}} and data \citep{vandermeer2023do_data} is available online.

\subsection{Data}
\label{sec:method-data}
We use \textbf{Debagreement} \citep{pougue2021debagreement} as the dataset containing  (dis-)agreement labels. This dataset contains user-submitted post pairs in English from five topics (Table~\ref{tab:data}), with post pairs annotated as \{agree, neutral, disagree\} by at least three crowd annotators.

We gather additional posts through the Reddit API using the usernames available in the Debagreement dataset. For each user still active, we collect up to 1000 most recent posts, which can be in any subreddit. The resulting posts range from September 2015 to April 2022. Subreddits host content on a variety of topics, not all of which encourage users to provide opinions based on their values. We are interested in finding preferences among values with respect to widespread societal issues, such as climate change. Thus, we filter out posts that are not likely to be of relevance to such issues. We
\begin{enumerate*}[label=(\arabic*)]
    \item exclude Not Safe For Work and entertainment-related subreddits, removing 1.4M posts,
    \item filter out noisy low-frequency subreddits (those with less than 50 collected posts), removing an additional 850K posts, and
    \item retain only English text posts, removing 377K posts.
\end{enumerate*}
Table~\ref{tab:data} shows the amount of data collected after filtering.

\begin{table}[!htb]
    \centering
    \begin{tabular}{@{}lccc@{}}
         \toprule
         \textbf{Subcorpus}         & \textbf{\# users} & \textbf{\# found} & \textbf{\# comments} \\
         \midrule
         \textsc{Brexit}            &  722 &  543 & 372K \\
         \textsc{Climate}           & 4580 & 3778 & 2.2M \\
         \textsc{BLM}               & 2516 & 2121 & 1.1M \\
         \textsc{Democrats}         & 6925 & 5646 & 3.8M \\
         \textsc{Republican}        & 8832 & 6839 & 3.9M \\
         \bottomrule
    \end{tabular}
    \caption{List of subcorpora gathered in Debagreement.}
    \label{tab:data}
\end{table}

\subsection{Value Extraction}
\label{sec:method-pred-val}
We formulate the value estimation task as recognizing whether a comment is related to a value by means of binary classification per value, matching the setup of \citet{qiu2021valuenet}. Our training data comprises general texts annotated for the presence of values across multiple domains. We combine data from two sources.
\begin{enumerate}[label=(\arabic*),nosep]
    \item \textbf{ValueNet} \citep{qiu2021valuenet}: We collapse non-neutral labels (1 and -1) into a single positive class and take the neutral labels (0) as a negative class. A non-neutral utility means that annotators considered the value to be relevant to the scenario, whereas the neutral class indicates that the value plays no apparent role.
    \item \textbf{ValueArg} \citep{kiesel2022identifying}: Their annotation scheme uses an updated (20) Schwartz values \citep{schwartz2012refining}, which we map back to the original 10 Schwartz values to allow joint training with the ValueNet dataset.
\end{enumerate}

We train all models with 10 seeds on random splits of learning data into train and validation sets to observe training stability. For both datasets, we split data into predefined learning (training and validation) and evaluation (test) sets. We ensure that all ten values occur equally frequently in the evaluation set. Each text sample is presented to our model ten times, once for each value by prepending a value-specific token. We describe the additional hyperparameters in the Appendix.

\subsection{Value Profile Estimation}
\label{sec:method-pred-profile}

Using a trained model, we construct a value profile $v$ per user by summing over value estimations of all individual messages. We assume relative frequencies of value mentions to be indicative of value preference similar to \citet{siebert2022estimating}.

To measure value conflicts, we introduce a lower limit $l$ on the total value mentions in each profile, i.e., requiring that each user has at least $l$ posts related to at least one value. Further, we normalize profile mention count by dividing it by the total number of value mentions per user. After this preprocessing, we compute the similarity $\mathcal{S}$ between two value profiles $v$ and $w$ in multiple ways.

\begin{description}[topsep=0pt, itemsep=0pt, leftmargin=0pt]
    \item [Kendall $\tau$] We sort value mentions by frequency and assign a rank label to each value. \mm{Kendall's rank correlation metric $\tau$ is a robust measure of correlation \citep{croux2010influence}, and considers the ranks of all pairs of values.} If a pair of values is ranked differently in $v$ than in $w$, the pair is considered discordant. Low scores indicate value conflict.
\begin{equation}
\label{eq:tau}
    \mathcal{S}^{\tau}(v,w) = 1 - \frac{2 \times (\text{\# discordant pairs)}}{\binom{n}{2}}
\end{equation}
    \item [Manhattan Distance (MD)] We compute the absolute difference between two profiles. High scores indicate value conflict.
\begin{equation}
    \mathcal{S}^{MD}(v,w) = \sum_{i=1}^{n} |v_i - w_i|
\end{equation}
    \item [Cosine (CO)] We compute traditional cosine similarity, low scores indicate conflict.
\begin{equation}
    \mathcal{S}^{CO}(v,w) = \frac{v\cdot w}{||v||\;||w||}
\end{equation}
    \item [Weighted-cosine (WC)] We compute a weighted cosine similarity that weighs similarities between values using the Schwartz Value Circumplex Model. For computing the similarity between value $v_i$ and $v_j$, we use a similarity matrix $\mathcal{B}$ constructed using a normal distribution with $\sigma=1$ centered on each value. Low scores indicate conflict.
\begin{equation}
    \mathcal{S}^{WC}(v,w) = \frac{\sum_{i=1}^{n} \mathcal{B}_i v_i w_i}{\sqrt{\sum_{i=1}^n \mathcal{B}_i v_{i}^2}\sqrt{\sum_{i=1}^n \mathcal{B}_i w_{i}^2}}
\end{equation}
\end{description}

\section{Experiments and Results}
We train models for value extraction and use those models to estimate value profiles. We check the consistency of our results with previous work, investigate differences in value profiles of disagreeing users, and perform qualitative analyses.

\begin{table}[!htb]
    \centering
    { \color{black}
    \small
    \begin{tabular}{@{}l@{\hspace{0.1cm}}c@{\hspace{0.1cm}}c@{\hspace{0.15cm}}c@{\hspace{0.15cm}}c@{}}
        \toprule
         \multirow{2}{*}{\textbf{Method}} & \multirow{2}{*}{\textbf{Training}} & \multicolumn{3}{c}{\textbf{Test}} \\
         \cmidrule{3-5}
         & & \textbf{ValueNet} & \textbf{ValueArg} & \textbf{Both} \\
        \midrule
         All-ones & -- & 0.40 & 0.11 & 0.26\\
         Value Dict. & -- & 0.45 & 0.64 & 0.57\\
         \citep{kiesel2022identifying}$^*$ & ValueArg & 0.15 & 0.37 & 0.28 \\
         \citep{qiu2021valuenet}$^*$ & ValueNet & 0.59 &  0.52  & 0.57\\
         \multirow[t]{3}{*}{BERT$_{\text{VE}}$} & ValueNet & \textbf{0.66} & 0.57 & 0.65 \\
         & ValueArg  & 0.46 & 0.76 & 0.67 \\
         & Both & 0.63 & \textbf{0.81} & \textbf{0.79} \\
         \multirow[t]{3}{*}{RoBERTa$_{\text{VE}}$} & ValueNet & 0.62 & 0.59 & 0.63 \\
         & ValueArg & 0.46 & 0.76 & 0.67 \\
         & Both & 0.63 & 0.78 & 0.78 \\
        \bottomrule
    \end{tabular}
    }
    \caption{Macro-averaged F$_1$ scores of the value estimation approaches on the value datasets. \mm{Methods marked with * are adapted for our comparison.}}
    \label{tab:value-cls-performance-condensed}
\end{table}
\subsection{Training Models for Value Estimation}
We experiment with two popular BERT-based models, BERT \citep{devlin2019bert} and RoBERTa \citep{liu2019text}, for value estimation. Further, we employ multiple baselines:
\begin{enumerate*}[label=(\arabic*)]
\item always predict all values for a comment (``All-ones''),
\item predict values based on mentions of value words from the \textbf{Schwartz Value Dictionary} \citep{ponizovskiy2020development},
\item the multi-label approach from \citet{kiesel2022identifying}, which uses an expanded label set, and
\item the utility model from \citet{qiu2021valuenet}.
\end{enumerate*}
\mm{The latter two baselines are BERT-based models. For \citet{kiesel2022identifying}, we use their multi-label setup to make predictions and map to the 10 Schwartz values at inference time (\emph{humility} and \emph{face} are not mapped to any value). Similarly, we map the rounded ternary utility labels from \citet{qiu2021valuenet} into binary value relevance labels at inference.}

\mm{Table~\ref{tab:value-cls-performance-condensed} shows the F$_1$ scores for the value extraction methods for different combinations of training and test datasets. We outperform all our baselines, including those from previous work. BERT$_{\text{VE}}$ and RoBERTa$_{\text{VE}}$ yield similar F$_1$ scores, and they perform best when trained on both datasets. We use our best-performing BERT$_{\text{VE}}$ model, trained on \emph{both} datasets, to construct the value profiles in the rest of the experiments.}

\subsection{Value Profile Estimation}

Table~\ref{tab:top-values} shows the top two frequent values in each domain. We observe that the distribution of values is specific to discussion contexts. For example, although stimulation is a common and frequent value, it is not the most frequent value in the \textsc{Brexit} subcorpus. We aggregate the values extracted for each user into their value profile. Table~\ref{tab:top-values} (last column) shows the mean pairwise $\tau$ distance (Equation~\ref{eq:tau}) among the value profiles in each domain. We observe that the \textsc{BLM} subcorpus has the most diversity among the five subcorpora.

\begin{table}[htb]
    \centering
    \begin{tabular}{@{}l@{\hspace{0.15cm}}l@{\hspace{0.15cm}}c@{}}
    \toprule
         \textbf{Subcorpus} & \textbf{Top Two Values} & \textbf{Avg. $\tau$}\\
    \midrule
         \textsc{Brexit}     & Security, Stimulation  & 0.260\\
         \textsc{climate}    & Stimulation, Security  & 0.308\\
         \textsc{BLM}        & Self-direction, Stimulation  & 0.343\\
         \textsc{democrats}  & Stimulation, Self-direction & 0.319\\
         \textsc{Republican} & Stimulation, Security & 0.315\\
    \bottomrule
    \end{tabular}
    \caption{Frequent values, and the mean similarity among value profiles in each domain.}
    \label{tab:top-values}
\end{table}

\begin{figure}[tb]
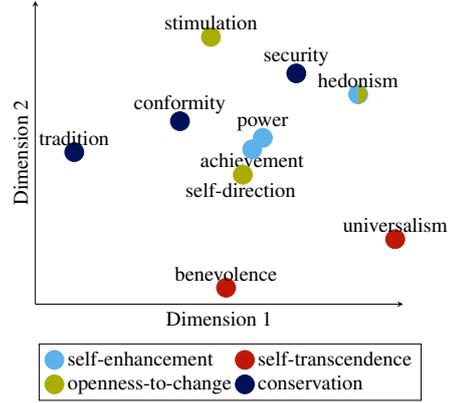

    \centering
    \includestandalone[width=0.8\columnwidth]{tikz/mds}
    \caption{Visualization of the covariance between values in estimated profiles.}
    \label{fig:mds}
\end{figure}

Next, to qualitatively assess the estimated value profiles, we normalize profiles (by the total number of value mentions) and compute covariance between profiles. Then, we perform
metric Multi-Dimensional Scaling (MDS) of the covariance matrix similar to \citet{ponizovskiy2020development}. Figure~\ref{fig:mds} shows a visualization of the first two dimensions after MDS. We observe that values that are close to each other in the Schwartz circumplex \citep{schwartz2012overview}, e.g., achievement and power, also tend to be closer in the MDS visualization.

\subsection{Value Conflicts and Disagreement}
We aim to analyze whether value conflicts influence disagreement in online discussions, using measurements of similarity between value profiles. We evaluate the following alternative hypothesis ($\mathbf{H_a}$) against a null hypothesis ($\mathbf{H_0}$).

\begin{description}[nosep]
\item [$\mathbf{H_0}$] The mean value profile similarity score between user pairs that disagree is equal to the mean value profile similarity score between user pairs that agree.
\item [$\mathbf{H_a}$] The mean value profile similarity score between user pairs that disagree is lower than the mean value profile similarity score between user pairs that agree.
\end{description}

We report the Bayes' Factor ($BF_{10}$)
\footnote{BF hypothesis tests are sensitive to the choice of prior. We use the implementation of \texttt{pingouin} \citep{vallat2018pingouin}, which includes a Jeffreys-Zellner-Siow prior, an objective prior for two-sample cases \citep{rouder2009bayesian}}
to assess the relative increase in odds for assuming the alternative over the null hypothesis after observing data \citep{azer2020not}. $BF_{10}$ scores in $[3^{-1},3]$ are considered to indicate evidence for neither hypothesis, whereas more extreme values favor one hypothesis over the other, allowing us to make conclusions in either direction \citep{kass1995bayes}.

We perform two experiments. First, we test the hypothesis for profiles constructed using the Value Profile Estimation (VPE) method. In the second experiment, we replace one of the profiles in each pair with a self-reported profile and agreement label. Thus, the second experiment removes some of the noise stemming from the VPE method.

\subsubsection{Profiles from VPE}
\label{sec:value-conflict-hypothesis}
We split Debagreement based on \emph{agree} and \emph{disagree} labels (and drop all pairs with a neutral label), obtaining respectively $G^{+}$ and $G^{-}$. For each group, we compute the profile similarity scores using each method mentioned in Section~\ref{sec:method-pred-val}. We do this per subreddit and observe the differences in score distributions. The alternative hypothesis is defined as the mean similarity scores in $G^{-}$ being lower\footnote{Higher for the \textbf{MD} metric, which flips the sign in Eqn.~\ref{eq:Ha}.} than the mean for $G^{+}$:
\begin{equation}
    \theta_G = \frac{1}{|G|}\sum_{\{p,c\} \in G} \mathcal{S}(p,c)
\end{equation}
\vspace{-1.5em}
\begin{align}
    H_0: \theta_{G^{-}} &= \theta_{G^{+}}\\
    H_a: \theta_{G^{-}} &< \theta_{G^{+}} \label{eq:Ha}
\end{align}

\begin{figure}[!htb]
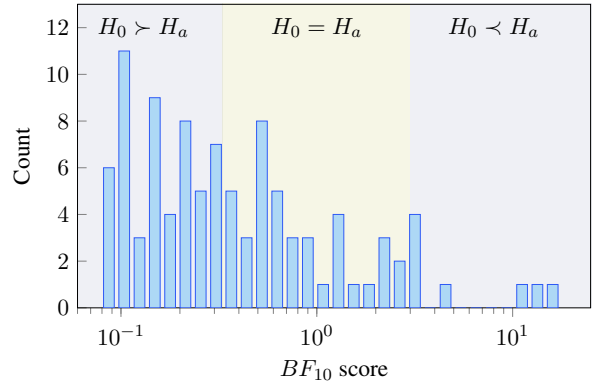

    \centering
    \includestandalone[width=\columnwidth]{tikz/bayesfactors}
    \caption{$BF_{10}$ scores obtained for the combinations of data, value estimation methods, and scoring metrics.}
    \label{fig:bf10-dist}
\end{figure}

We report the $BF_{10}$ for all combinations of similarity methods and parameters. We run 100 tests, considering 5 subreddits, 4 similarity scores, and 5 value profile thresholds $l=\{1,10,50,200,500\}$. Figure~\ref{fig:bf10-dist} provides an overview of the $BF_{10}$ scores.

First, we observe that a majority of the combinations show stronger support for accepting the null hypothesis over the alternative hypothesis (i.e., most scores fall inside the leftmost blue bin). This indicates that value conflicts may not be directly correlated to disagreement in many cases. Possibly, other content-related factors play a stronger role in these discussions. However, there are some tests that still show evidence for rejecting the null hypothesis ($BF_{10} > 3$).

Thus, given specific settings and domains, we can trace disagreement between users to value conflicts. Table~\ref{tab:top-5-vpe-vpe} shows the tests where $BF_{10} > 3$. In all cases, the filter $l$ was 10 or more, stipulating that populated value profiles are required for measuring value conflicts reliably.
We observe that \textsc{BLM}, the subcorpus with the highest profile diversity (Table~\ref{tab:top-values}), is frequent among these positive cases. Thus, having diverse profiles increases the likelihood of finding a link between values and disagreement.
One positive test result is observed for the \textsc{Brexit} subcorpus for a high profile threshold (500). Brexit includes the smallest number of user profiles; the high profile threshold further removes several profiles. Thus, the positive result for \textsc{Brexit}, based on a low number of profile comparisons, may not be reliable.

\begin{table}[tb]
    \centering
    {\color{black}
    \begin{tabular}{@{}p{0.7cm}P{2cm}P{1.6cm}P{1.6cm}@{}}
    \toprule
         \textbf{$BF_{10}$} & \textbf{Subreddit} & \textbf{Similarity score} & \textbf{Profile threshold}  \\
    \midrule
        17.451 & \textsc{BLM} & CO & 10 \\
        12.485 & \textsc{BLM} & WC & 10 \\
        10.504 & \textsc{BLM} & $\tau$ & 250 \\
        4.223 & \textsc{BLM} & MD & 10 \\
        3.442 & \textsc{Brexit} & WC & 500\\
        3.159 & \textsc{BLM} & WC & 50 \\
    \bottomrule
    \end{tabular}
    }
    \caption{The six tests between two VPE-constructed profiles with $BF_{10} > 3$.}
    \label{tab:top-5-vpe-vpe}
\end{table}

\subsubsection{Mixing with Self-reported Profiles}
\label{sec:value-conflict-hypothesis-human}
Given that we use a novel method for estimating value profiles, we compare the results from the previous experiment with one that uses self-reported value profiles. Self-reported profiles mitigate the noise stemming from the value estimation step. The setup is identical to Section~\ref{sec:value-conflict-hypothesis}, but now we compute similarities between an estimated profile and a self-reported profile, obtained from a value survey.

We run a user study to obtain \begin{enumerate*}[label=(\arabic*)]
    \item self-reports of value profiles using an established value survey \citep[PVQ-21,][]{schwartz2021repository}, and
    \item agreement labels on posts in Debagreement.
\end{enumerate*}
We obtained an IRB approval (exempt status) for our study.

We collected annotations from 26 Prolific (\url{prolific.co}) users. We selected five task instances for each subreddit from Debagreement posts with populated value profiles, rendering testing on multiple profile thresholds unnecessary. We removed three task instances, which obtained a majority of neutral and not-enough-information judgments, leaving 22 rated instances. Thus, our analyses include a total of 572 judgments.

\begin{figure}[tb]
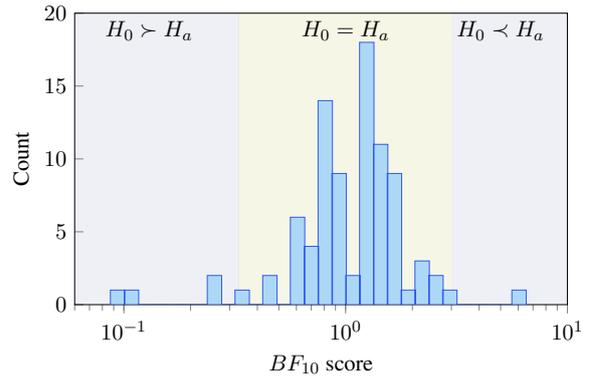

    \centering
    \includestandalone[width=\columnwidth]{tikz/bayesfactors_human}
    \caption{$BF_{10}$ scores for all similarity scores and task instances comparing VPE and self-reported profiles.}
    \label{fig:bayes_human}
\end{figure}

The results are shown in Figure~\ref{fig:bayes_human}. We observe that deciding between the two hypotheses is not possible, in a majority of cases, as most evidence attributed both as equally likely. However, it is interesting to notice that using self-reported value profiles shifts the majority of results from favoring the null hypothesis to the undecidable range. In combination with the results from the previous section, this indicates that VPE methods need careful evaluation with respect to self-reported profiles as both may contain errors stemming from different sources and may have complementary merits. VPE suffers from errors made by the value estimation model but has the potential to use large amounts of data. In contrast, although self-reports yield a profile directly, they may be prone to biases.

Two tests still show evidence in favor of accepting $H_a$ (see Table~\ref{tab:top-5-vpe-human}). They are on two task instances in the same domain, \textsc{democrats}, and are measured for the $\tau$ and $MD$ metrics.
Here, our results differ from the previous experiment, and different subreddits result in high $BF_{10}$ scores. In this case, one user's value profile is constructed using self-reports, which are obtained without reference to discussions (i.e. not estimated from posts on Reddit). This may cause other factors to influence the diversity of profiles stemming from the PVQ.
Furthermore, the task instances contained a call for action (e.g., \emph{Please just vote [..]} and \emph{The gloves should come off [..]}). The values embedded in the call to action may be one of the reasons why annotators felt inclined to disagree or agree.

\begin{table}[ht]
    \centering
    {\color{black}
    \begin{tabular}{@{}p{0.7cm}P{2cm}P{3cm}@{}}
    \toprule
         \textbf{$BF_{10}$} & \textbf{Subreddit} & \textbf{Similarity score} \\
    \midrule
        6.490 & \textsc{democrats} & $\tau$ \\
        3.066 & \textsc{democrats} & MD \\
        \midrule
        2.543 & \textsc{Brexit} & MD \\
        2.407 & \textsc{Brexit} & CO \\
        2.230 & \textsc{climate} & CO\\
    \bottomrule
    \end{tabular}
    }
    \caption{The top-five $BF_{10}$ scores, when comparing a VPE-constructed profile and a self-reported profile.}
    \label{tab:top-5-vpe-human}
\end{table}

\subsubsection{Qualitative Assessment}
To better understand when value conflicts influence disagreement, we perform a qualitative analysis of some instances (comment pairs) from the dataset that follow our hypothesis and some that do not (Figure~\ref{app:fig:confusion-matrix} in Appendix~\ref{app:additional-results} shows such examples).

We identify five trends in misaligned instances.
\begin{enumerate*}[nosep, leftmargin=*, label=(\arabic*)]
    \item \textbf{Not enough information} in a value profile (i.e., low-frequency value mentions). This means that the user posted little value-laden content or that the value extraction method erroneously ignored some value-laden comments.

    \item \textbf{No apparent value-based reasoning} involved in the comments, e.g., factual answers to a question.

    \item \textbf{(Dis-)agreement} happens \textbf{on a content level} since profiles do not dictate individual utterances. This occurs when users disagree that a decision is ``for good,'' but fail to motivate their motivations for what is ``good.''

    \item The \textbf{target} of disagreement can be \textbf{partial}, whereas value conflicts are measured between two users.

    \item In a few cases, the label given in Debagreement is \textbf{faulty} (e.g., annotators misinterpreting sarcasm or the text is vague).
\end{enumerate*}

\subsection{Use Case: Predicting (Dis-)agreement}
\label{sec:method-pred-arg}
We assume that users' value profiles (in addition to the content of users' posts) play a role in predicting the agreement between users. 
We adopt the setup from \citet{pougue2021debagreement}, where an agreement label is predicted between parent $p$ and child comments $c$. We add extra information to $p$ and $c$ using four methods.
\begin{description}[nosep, leftmargin=1em]
    \item [Random noise ($\epsilon$)] Random noise to test for spurious correlations.
    \item [User centroids ($z$)] Centroids of all posts from a single user by constructing TF-IDF vectors for each post and then taking an average.
    \item [Explicit user features ($u$)] Nine features commonly extracted for representing users on Reddit (e.g., \citep{jhaver2019does, chew2021predicting}) to add extra contextual information. 
    \item [Value profile ($v$)] Value estimation on user posts to extract an explicit value profile for the ten Schwartz values.
\end{description}
We create embeddings (TF-IDF or BERT) for $p$ and $c$ and concatenate them to the user-specific context \citep{gu2021package}. We standardize the user-specific context information to avoid raw values having a large impact, similar to the value profiles ($v$). When training with user profiles, we subsample Debagreement to include only those $(p,c)$ pairs in which we have background data for both $p$ and $c$. This leaves 65\% of the data (28K samples). We train our classifier on an 80/10/10 split, retaining the most recent $20$\% as validation and test sets to reflect a real-world training scenario on historical data \citep{sogaard2021we}.
\begin{figure}[tb]
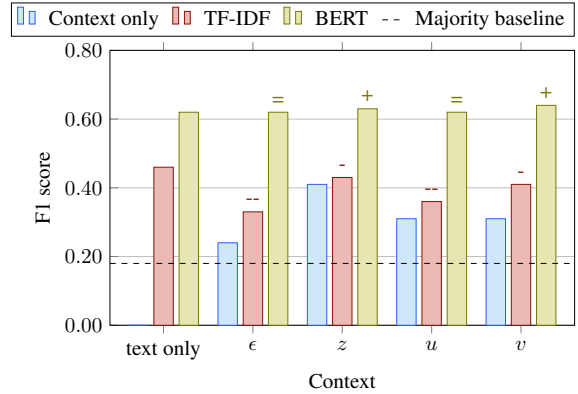

    \centering
    \includestandalone[width=\columnwidth]{tikz/use_case_agreement}
    \caption{$F_1$ scores when adding extra context information. Symbols above bars show changes with respect to text-only: $--$ for $\Delta F_1 < -0.1$; $-$ for $-0.1 < \Delta F_1 < 0$; $=$ for $\Delta F_1 = 0$; and $+$ for $\Delta F_1 > 0$.}
    \label{fig:use-case-change}
\end{figure}

Figure~\ref{fig:use-case-change} shows the results. Classifiers using TF-IDF embeddings fail to use the information effectively.
BERT outperforms both our baselines, in line with the results for \citep{pougue2021debagreement}. In this setting, none of the additional information causes major changes in performance, but we see an improvement using the value profiles and centroids. Compared to other work, using user-specific information is surprisingly difficult \citep{al2020exploiting}. Further inspection for BERT indicates that the \emph{neutral} class is hard to predict, as
information from the value profiles may not be relevant. Mixing background information using, e.g., GNNs \citep{luo2023improving} may make more effective use of the profile information.

\section{Conclusion}
Our results on the role of value conflicts in disagreements are mixed.
On the one hand, we mostly note negative evidence of a correlation between profile similarity and disagreeing users when using the VPE methods. When using self-reported profiles, the negative evidence reduces and results become inconclusive for a majority of the cases. This suggests that the nature of the profiles differs, and further investigation is necessary.

On the other hand, we observe that value conflicts were found to lead to disagreements in specific cases. When values are likely to be relevant and diverse, we find evidence for a correlation between value conflict and disagreement. While value conflicts may not be directly related to disagreement, they do signal diversity with respect to the underlying motivations of participants.

Using value profiles in combination with BERT performs marginally better than a text-only baseline in predicting agreement. Yet, VPE can be valuable for characterizing and enhancing diversity in discussions. Further, making participants value-aware could enhance the discussion quality.

Constructing profiles from behavioral cues, such as written opinions, is noisy. For future work, we hope to see the creation of resources that allow end-to-end evaluation by combining text posts with a consistent set of users that allows aggregation to ground truth profiles or self-reported profiles. However, gathering such profile information outside controlled lab settings is highly complex. Future experiments may incorporate more judgments and provide stronger evidence for one hypothesis. These can be retrofitted with our results through Bayesian updating \citep{moerbeek2021bayesian}.

\section*{Limitations}
We outline four limitations of our work related to the experimental setup and the interpretation of results that are specific to the modeling of value conflicts in online discussions.

First, the value extraction methods we employ (see Table 2) may have unknown errors. Our work is not focused on optimizing value extraction, which is an emerging research direction \citep{kiesel2023semeval}. Adding more annotated Reddit data would allow us to judge the performance of value extraction models better. A future direction is to employ other training paradigms like Multi-task Learning \citep[e.g.,][]{fang2019neural} or techniques for mixing in general-purpose language models \citep[e.g.,][]{vandermeer2022will}.

Second, we obtain the self-reported value profiles with the PVQ-21 questionnaire (see Section 4.4). Since we run the questionnaire before starting an annotation experiment to obtain agreement labels, there may be ordering bias in the obtained labels. The experiments could be enhanced by swapping the order of PVQ-21 and the annotation tasks to estimate the effect of answering the questionnaire on the agreement labels.

Third, the reporting of our results is limited to the Bayes Factor (BF). Further, most of our results fall inside the neutral category (``cannot decide between $H_0$ and $H_a$''). We require more data to decide which of the hypotheses is more likely. An estimation of the posterior odds of the hypotheses e.g., in the form of \emph{Highest Density Intervals} (HDI) might yield more insights, and would involve deciding on a \emph{region of practical equivalence} (ROPE), as well as picking a thus far unknown prior distribution over the values for $\mathcal{S}$ in our two hypotheses \citep{kruschke2018rejecting}. However, BF and HDI interpretations can be seen as complementary, respectively quantifying evidence or beliefs \citep{van2022advantages}.

Lastly, our qualitative findings are derived from examining online interactions with limited context. To obtain a more complete picture, both the values and the interpretation of the author's role in discussions should be verified by the authors themselves. However, running such experiments in controlled lab settings is beyond the scope of our work since we focus on disagreements in online discussions.

\section*{Ethics Statement}
First, the dataset used to model online discussions, Debagreement, was sourced from online interactions between users on Reddit. Research conducted on Reddit data is biased to a WEIRD (\underline{W}estern, \underline{E}ducated, \underline{I}ndustrialized, \underline{R}ich, \underline{D}emocratic) demographic, and results may not generalize to a broader set of users \citep{proferes2021studying}. However, our method outlines which data is required for performing the same analysis given the availability of richer data, not necessarily stemming from Reddit.
Second, models for predicting values may be wrong, they may lead to harmful outcomes for particular groups or populations \citep{mehrabi2021survey}. In any application, the incorporation of control mechanisms (i.e., providing users a way to influence the construction of their own value profile) is a requirement for making sure the value profiling is conducted in a transparent and accountable manner. Broadly, this work should further be situated in a system containing checks and balances, making sure any output stemming from automated classification is verified by human agents before having an effect on actual users.

\section*{Acknowledgements}
This research was funded by the Netherlands Organisation for Scientific Research (NWO) through the Hybrid Intelligence Centre via the Zwaartekracht grant (024.004.022). We would like to thank the ARR reviewers for their feedback.

\bibliography{papers, extra}

\clearpage
\appendix
\section{Methodological details}
\subsection{Training Value extraction methods}
For training our Transformer-based NLP models, we turned to the Huggingface \texttt{transformers} Python package \citep{wolf2020transformers}. See Table~\ref{app:tab:hyperparams-value-extraction} for the hyperparameters used for training value extraction models. All computational experiments were run on machines containing up to 2x 3090 Nvidia RTX GPUs. Training a single value extraction model takes around 3 hours. Running VPE on background data takes significantly longer due to the number of inferences made, up to 7 days of computation.

\begin{table}[ht]
    \centering
    \begin{tabular}{ll}
    \toprule
         \textbf{Hyperparameter} & \textbf{Value}  \\
    \midrule
         \texttt{train epochs} & 10\\
         \texttt{learning rate} & $5e-05$\\
         \texttt{model} & bert-base-uncased\\
         \texttt{batch size} & 256\\
    \bottomrule
    \end{tabular}
    \caption{Hyperparameters used for training models for value extraction}
    \label{app:tab:hyperparams-value-extraction}
\end{table}

\paragraph{Filtering Reddit data}
We construct value profiles from the data scraped from Reddit, from which we filter posts not likely to be of relevance to discussing widespread societal issues. We remove posts from \begin{enumerate*}[label=(\arabic*)]
\item NSFW subreddits\footnote{\url{https://www.reddit.com/r/ListOfSubreddits/wiki/nsfw}},
\item gaming subreddits\footnote{\url{https://www.reddit.com/r/gaming/wiki/faq}},
\item image-related subreddits\footnote{\url{https://www.reddit.com/r/ListOfSubreddits/wiki/sfwporn}},
\item user subreddits, all subreddits starting with ``\texttt{u\_}'',
\item non-English posts (as detected using the FastText \citep{joulin2017bag} Language Identification model\footnote{\url{https://fasttext.cc/docs/en/language-identification.html}}),
\item and subreddits for which we could extract less than 50 posts.
\end{enumerate*}

\paragraph{Using Value Dictionary for VPE}
We use the following pipeline for constructing value profiles using the \textbf{Schwartz Value Dictionary}.
\begin{enumerate}
    \item Load words from \citet{ponizovskiy2020development}. Some values have more words in the dictionary, and thus we introduce a weighting scheme to normalize over the number of words, such that a value $v$ inside the profile with relatively few dictionary words has a higher weight $w_v$.
    \item Replace URLs with a special \texttt{[URL]} token.
    \item Apply lemmatization to all comments from a single user.
    \item Classify individual comments for values. If a comment contains at least one term from the VD, classify the comment as being relevant for that value.
    \item Aggregate over all comments.
    \item Apply weighting $z = count(v) \times w_v$.
    \item Apply normalization over the profile so it sums to 1.
\end{enumerate}

\subsection{Annotator experiment}
We separated our annotator experiment into two phases: \begin{enumerate*}[label=(\arabic*)]
    \item the filling in of the PVQ-21, and
    \item providing judgments on posts from Debagreement.
\end{enumerate*}
The first phase was performed through Qualtrics questionnaire software. We provide screenshots of all steps (informed consent, annotation instructions) below. The second phase is hosted on Prodigy \citep{prodigy_montani_honnibal}.

\begin{figure*}[!htb]
    \centering
    \includegraphics[width=\textwidth]{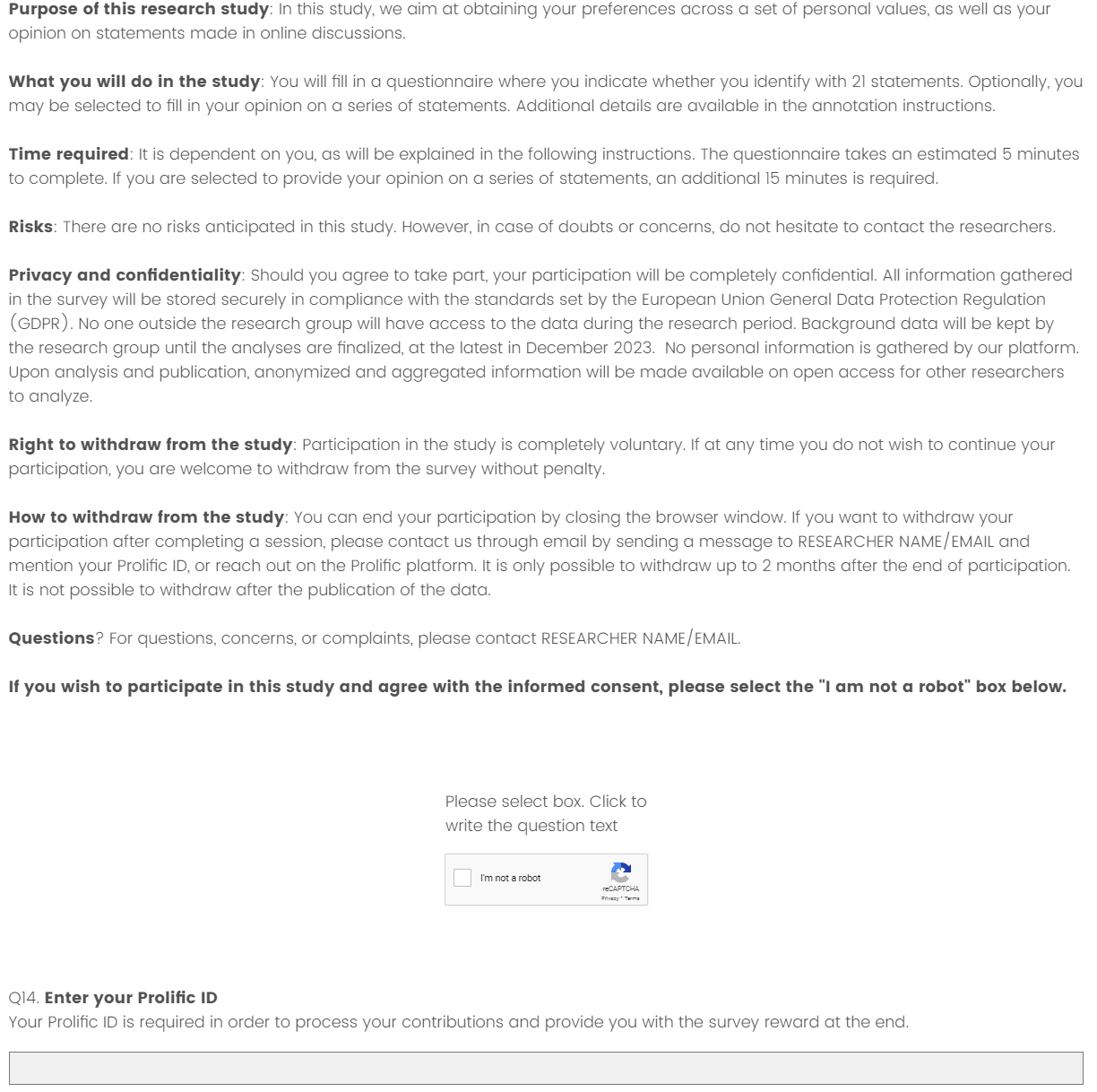}
    \caption{Informed consent shown to users before starting the experiment.}
    \label{app:fig:informed-consent}
\end{figure*}

\begin{figure*}[!htb]
    \centering
    \includegraphics[width=\textwidth]{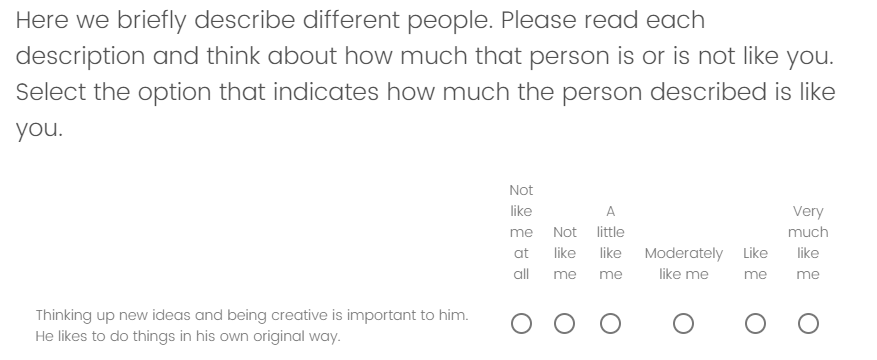}
    \caption{Screenshot of the PVQ-21 survey.}
    \label{app:fig:qualtrics-screenshot}
\end{figure*}

\begin{figure*}[!htb]
    \centering
    \includegraphics[width=\textwidth]{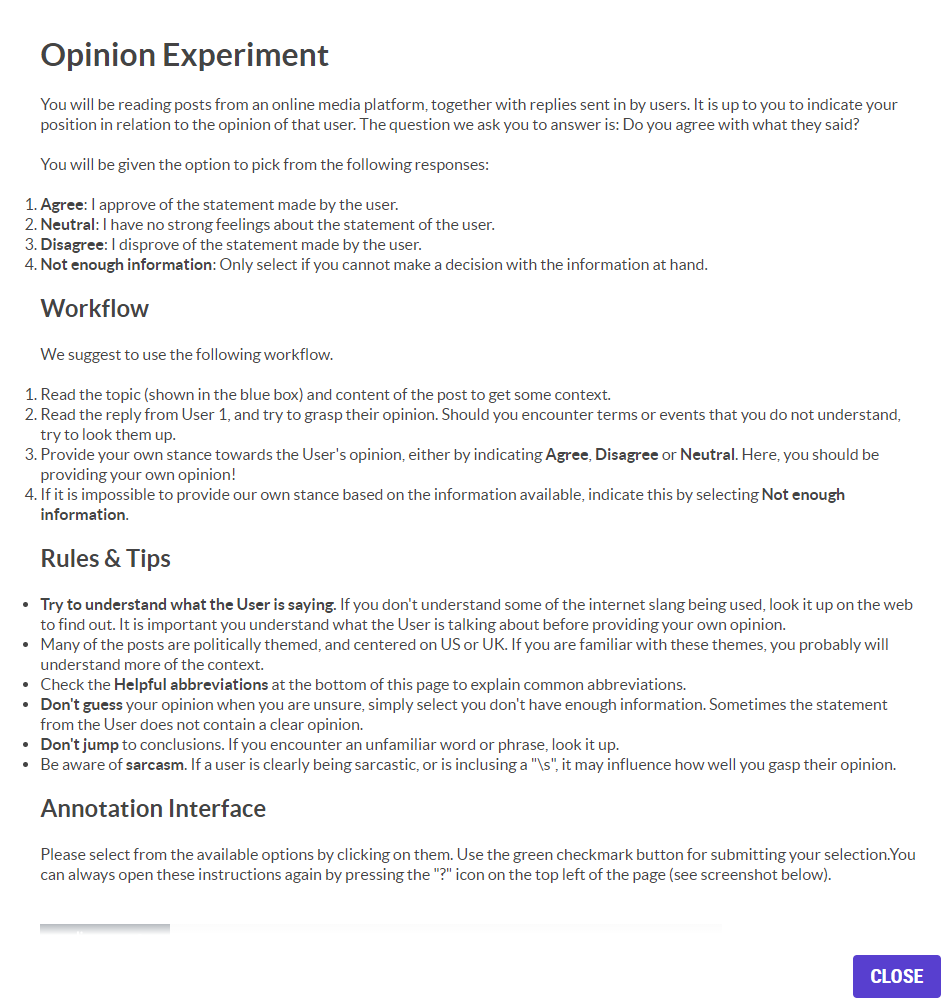}
    \caption{Instructions shown to users for the annotation experiment.}
    \label{app:fig:prodigy-instructions}
\end{figure*}

\begin{figure*}[!htb]
    \centering
    \includegraphics[width=\textwidth]{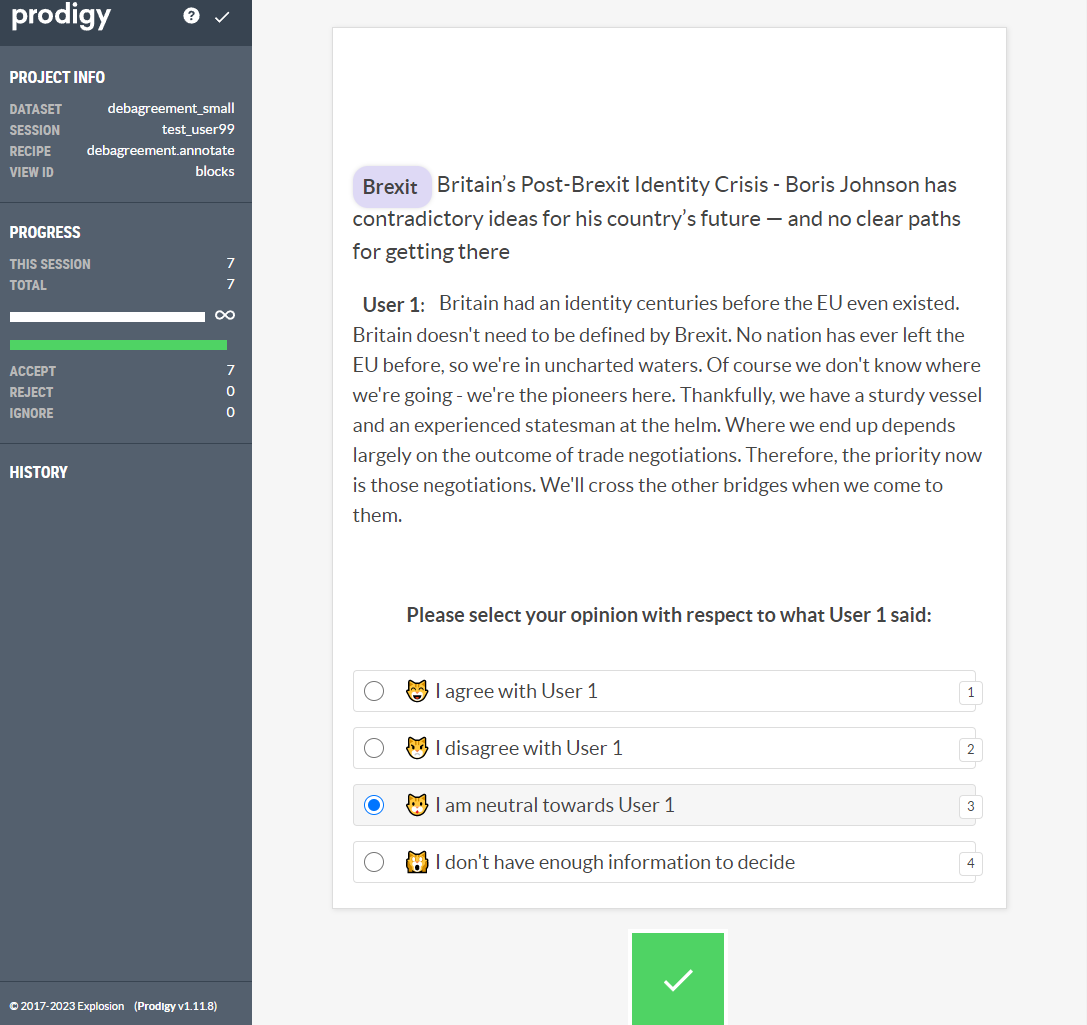}
    \caption{Annotation interface.}
    \label{app:fig:prodigy-interface}
\end{figure*}

\begin{itemize}
    \item \textbf{Informed consent} See Figure~\ref{app:fig:informed-consent}. Shown to users before starting the experiment outlining the data protection and disclaimers of any risks.
    \item \textbf{Value Survey} See Figure~\ref{app:fig:qualtrics-screenshot}. Users fill in 21 items on a Likert scale.
    \item \textbf{Annotation instructions} See Figure~\ref{app:fig:prodigy-instructions}.
    \item \textbf{Annotation interface} See Figure~\ref{app:fig:prodigy-interface}. Users were asked to fill in 25 task instances (five per subcorpus) on the annotation platform.
\end{itemize}

Annotators were recruited from the Prolific (\url{prolific.co}) crowd worker platform. All participants were paid at least the recommended \pounds 9/h wage, and on average spent 20 minutes on the two tasks combined. This payment is considered an ethical reward according to Prolific.

\paragraph{Transforming survey responses into profiles}
We adopt the suggestions from \citet{schwartz2012overview} for constructing a numerical value profile that reflects preferences among values. We create the following pipeline:
\begin{enumerate}
    \item Gather Likert-scale answers on all 21 items.
    \item Check if two attention check items were correctly answered. Participants were asked to fill in a given score. Disregard participant results otherwise.
    \item Compute \underline{M}ean \underline{Rat}ing for each participant (MRAT).
    \item Subtract the mean score from all other scores to obtain centered response scores.
    \item Normalize the profile by dividing by the sum of all scores.
\end{enumerate}

\subsection{Training agreement analysis models}
Training models for agreement analysis takes around 4 hours for the BERT models on the subsampled Debagreement dataset. See Table~\ref{app:tab:hyperparams-agreement-analysis} for the hyperparameters used. Debagreement may be reused under the CC BY 4.0 license. For the implementation of the TF-IDF, we used the \texttt{sklearn} \citep{scikit-learn} Python package. All training involving TF-IDF embeddings takes under 1 hour.
\begin{table}[!htb]
    \centering
    \begin{tabular}{lc}
    \toprule
         \textbf{Hyperparameter} & \textbf{Value}  \\
    \midrule
         \texttt{train epochs} & 7\\
         \texttt{learning rate} & $5e-05$\\
         \texttt{model} & bert-base-uncased\\
         \texttt{batch size} & 64 \\
    \bottomrule
    \end{tabular}
    \caption{Hyperparameters used for training models for agreement analysis}
    \label{app:tab:hyperparams-agreement-analysis}
\end{table}

We constructed three types of extra user information for the agreement analysis task:
\begin{description}[topsep=0pt, itemsep=0pt, leftmargin=0pt]
    \item [Random noise] We sample a vector of size 768 from a random uniform distribution over $[0,1)$.
    \item [User centroids] We stem the posts from users that contain at least one value term according to the value dictionary and transform comments to TF-IDF vectors. We restrict the vocabulary to the 768 most frequent terms. We then compute the average over all vectors for a single user.
    \item [Explicit user features] We construct user feature vectors for Reddit users through the Reddit PRAW API. See Table~\ref{app:tab:reddit-user-features} for the features used.
\end{description}

\begin{table}[!htb]
    \centering
    \begin{tabular}{p{2.3cm}p{4.7cm}}
    \toprule
         \textbf{Feature} & \textbf{Explanation}  \\
    \midrule
         comment\_karma     & Total amount of upvotes minus downvotes on comments.\\
         link\_karma        & Total amount of upvotes minus downvotes on link submissions.\\
         date\_created      & Timestamp of account creation.\\
         gold\_status       & Whether the user is a gold member.\\
         mod\_status        & Whether the user is a mod of any subreddit.\\
         employee\_status   & Whether the user is an employee of Reddit.\\
         num\_gilded        & Number of gilded items.\\
         num\_comments      & Number of comments posted by user.\\
         num\_links         & Number of links submitted by user.\\
    \bottomrule
    \end{tabular}
    \caption{Features used to represent a user from Reddit} 
    \label{app:tab:reddit-user-features}
\end{table}

\section{Additional Results}
\label{app:additional-results}
\subsection{Value Extraction}
For a complete overview of the performance of the value extraction models, including the standard deviation over 10 random seeds for the $VE$ models, see Table~\ref{tab:value-cls-performance}.
\begin{table*}[!htb]
    \centering
    {\color{black}
    \small
    \begin{tabular}{@{}lcccccccc@{}}
        \toprule
         \textbf{Method} & \makecell[c]{\textbf{Training}\\ \textbf{data}} & \textbf{P(VN)} & \textbf{R(VN)} & \textbf{F1(VN)}& \textbf{P(VA)} & \textbf{R(VA)} & \textbf{F1(VA)} & \textbf{F1}\\
        \midrule
         All-ones           & --            & 0.34 & 0.50 & 0.40 & 0.11 & 0.50 & 0.18 & 0.26\\
         VD                 & --            & 0.56 & 0.55 & 0.45 & 0.64 & 0.58 & 0.59 & 0.57\\
         \citep{kiesel2022identifying}$^*$  & \emph{VA} & 0.20 & 0.21 & 0.15 & 0.47 & 0.34 & 0.37 & 0.28\\
         \citep{qiu2021valuenet}$^*$        & \emph{VN} & 0.64 & 0.65 & 0.59 & 0.53 & 0.52 & 0.52 & 0.57\\
         BERT               & \emph{VN}     & $0.66_{\pm0.00}$& $0.68_{\pm0.00}$& $0.66_{\pm0.00}$& $0.57_{\pm0.02}$& $0.60_{\pm0.02}$& $0.57_{\pm0.03}$& $0.65_{\pm0.02}$\\
                            & \emph{VA}     & $0.57_{\pm0.00}$& $0.56_{\pm0.00}$& $0.46_{\pm0.00}$& $0.79_{\pm0.02}$& $0.74_{\pm0.01}$& $0.76_{\pm0.01}$& $0.67_{\pm0.01}$\\
                            & \emph{Both}   & $0.63_{\pm0.00}$& $0.64_{\pm0.00}$& $0.63_{\pm0.00}$& $0.84_{\pm0.02}$& $0.79_{\pm0.00}$& $0.81_{\pm0.01}$& $0.79_{\pm0.00}$\\
         RoBERTa            & \emph{VN}     & $0.61_{\pm0.15}$& $0.66_{\pm0.05}$& $0.62_{\pm0.12}$& $0.58_{\pm0.02}$& $0.61_{\pm0.02}$& $0.59_{\pm0.02}$& $0.63_{\pm0.03}$\\
                            & \emph{VA}     & $0.57_{\pm0.00}$& $0.56_{\pm0.00}$& $0.46_{\pm0.00}$& $0.79_{\pm0.02}$& $0.74_{\pm0.01}$& $0.76_{\pm0.01}$& $0.67_{\pm0.01}$\\
                            & \emph{Both}   & $0.63_{\pm0.00}$& $0.64_{\pm0.00}$& $0.63_{\pm0.00}$& $0.83_{\pm0.02}$& $0.78_{\pm0.01}$& $0.80_{\pm0.01}$& $0.78_{\pm0.00}$\\
        \bottomrule
    \end{tabular}
    }
    \caption{Macro-averaged performance of the value estimation approaches on the value datasets, showing averages and standard deviation for our own models over 10 different seeds. \mm{\emph{VN} denotes ValueNet, \emph{VA} denotes ValueArg. Methods marked with * are trained on a different objective than our VE task.}}
    \label{tab:value-cls-performance}
\end{table*}


\subsection{Value Survey}
\paragraph{Demographics}
We received a total of 27 responses, one of which was ignored because of a failed attention check. Different ages were represented in our sample (M=28.0, SD=8.7), and annotators originated from Europe (18 annotators), South Africa (8 annotators), the UK (1), and the US (1). About half (13) were registered students.

\paragraph{Reliability}
Since the PVQ has two questions for each personal value, we are able to compute internal consistency using Cronbach $\alpha$ per value. See the results in Table~\ref{app:tab:cronbach}. We observe a wide range of reliability scores, of which only conformity reaches above a score of 0.7. Most interestingly, we see that tradition is of very low reliability, possibly due to the demographic of some of our participants (students). Three task instances received mostly neutral or not-enough-information labels, and were disregarded in our analysis.
\begin{table}[!htb]
    \centering
    \begin{tabular}{@{}lcc@{}}
        \toprule
         \textbf{Value }& \textbf{$\alpha$} & \textbf{95\% CI}\\
        \midrule
         conformity     & 0.717 & (0.514,0.835) \\
         tradition      & 0.051 & (-0.627,0.447) \\
         benevolence    & 0.336 & (-0.138,0.613) \\
         universalism   & 0.407 & (-0.016,0.654) \\
         self-direction & 0.641 & (0.384,0.790) \\
         stimulation    & 0.589 & (0.295,0.760) \\
         hedonism       & 0.618 & (0.345,0.777) \\
         achievement    & 0.504 & (0.149,0.711) \\
         power          & 0.371 & (-0.078,0.633) \\
         security       & 0.388 & (-0.050,0.643) \\
        \bottomrule
    \end{tabular}
    \caption{Internal consistency scores (Cronbach's $\alpha$) for the values in the PVQ-21 questionnaire.}
    \label{app:tab:cronbach}
\end{table}

\subsection{Qualitative Examples of Value Conflicts and (Dis-)agreement}

\begin{table*}[!htb]
    \centering
    \begin{tabular}{ccc}
    \toprule
    & \textbf{Disagree} & \textbf{Agree} \\
    \midrule
    \parbox[t]{2mm}{\rotatebox[origin=l]{90}{\textbf{No Value Conflict}}} &
    \begin{tikzpicture}
        \draw[fill=blue!10](0,0) rectangle (7,1.25) node[midway,align=left](m1){This is NOT a public statue. It's a privately\\ owned statue on private property.. the\\ government has zero right to take it down.};
        \draw[fill=red!30](1,-0.25) rectangle (7,-1.75) node[midway,align=left](m2){Not so sure. A crime on private\\ property is still a crime, and\\ defending racism is a crime.};
        \begin{scope}[on background layer]
            \draw (m1.west) |- (m2.west);
        \end{scope}
    \end{tikzpicture}  &
    \begin{tikzpicture}
        \draw[fill=blue!10](0,0) rectangle (7,1.25) node[midway,align=left](m1){Climate justice has waited too long to be \\served. The time is now!};
        \draw[fill=blue!30](0.33,-0.25) rectangle (7,-2.25) node[midway,align=left](m2){Guys, get out there and support people,\\ politicians, businesses, companies, and\\ local stores who support climate justice\\ and sustained efforts to promote\\ sustainability and eco-friendliness alike!!};
        \begin{scope}[on background layer]
            \draw (m1.west) |- (m2.west);
        \end{scope}
    \end{tikzpicture} \\
    \midrule
    \parbox[t]{2mm}{\rotatebox[origin=l]{90}{\textbf{Value Conflict}}} &
    \begin{tikzpicture}
        \draw[fill=blue!10](0,0) rectangle (7,1.25) node[midway,align=left](m1){The EU moves very slowly.. Don’t blame\\ the UK if the EU is so slow.};
        \draw[fill=red!30](1,-0.25) rectangle (7,-1.75) node[midway,align=left](m2){So you're saying the EU should make \\the UK its priority? Why should the \\UK have priority over another issue?};

        \begin{scope}[on background layer]
            \draw (m1.west) |- (m2.west);
        \end{scope}
    \end{tikzpicture}  &
    \begin{tikzpicture}
        \draw[fill=blue!10](-0.25,0) rectangle (7,1.25) node[midway,align=left](m1){Brexit is a symptom, not a problem in itself.\\ Don't just make the symptom go away,\\ treat the many underlying problems first};
        \draw[fill=blue!30](0,-0.25) rectangle (7,-1.75) node[midway,align=left](m2){I agree, but you have a parliament that\\ took control from May then did the \\dumbest thing it could do by not voting \\for any of the proposals.};

        \begin{scope}[on background layer]
            \draw (m1.west) |- (m2.west);
        \end{scope}
    \end{tikzpicture} \\
    \bottomrule
    \end{tabular}
    \caption{Confusion matrix of qualitative examples of the match between value conflict and (dis-)agreement.}
    \label{app:fig:confusion-matrix}
\end{table*}

We perform a qualitative analysis of some instances (comment pairs) from the dataset that follow our hypothesis and some that do not to gain a better understanding of when value conflicts influence disagreement. Table~\ref{app:fig:confusion-matrix} shows examples of the types of pairs we analyze.

\subsection{Decomposition of \texorpdfstring{$BF_{10}$}{BF10} results}

\begin{figure}[!htb]
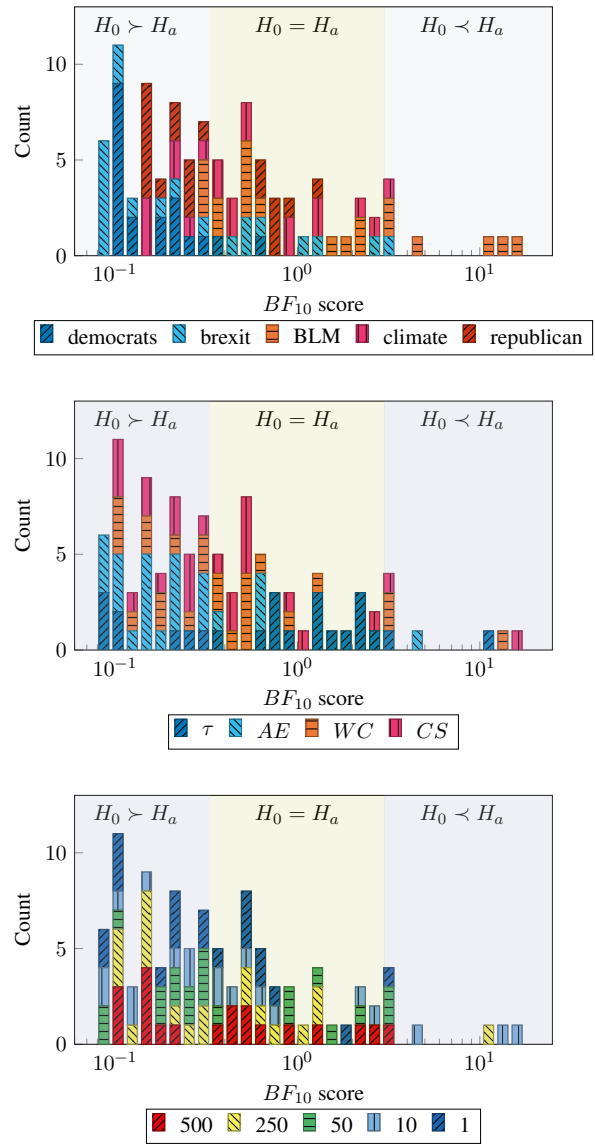

    \centering
    \includestandalone[width=\columnwidth]{tikz/BF_colored}
    \caption{$BF_{10}$ scores when testing between two VPE-constructed profiles, obtained for all combinations of subreddits (top figure), similarity scores (middle figure) and profile thresholds (bottom figure).}
    \label{fig:bf10-decomp}
\end{figure}

\begin{figure}[!htb]
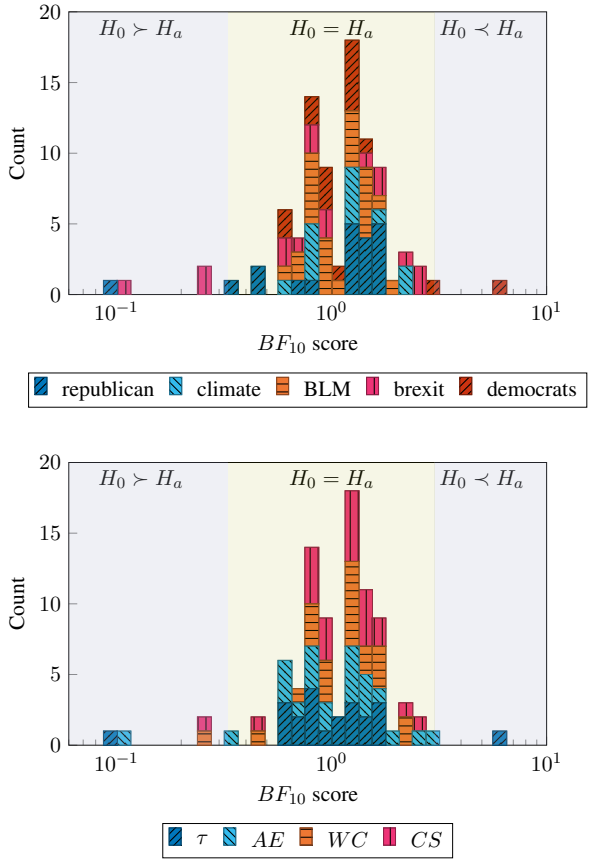

    \centering
    \includestandalone[width=\columnwidth]{tikz/BF_human_colored}
    \caption{$BF_{10}$ scores when testing between a VPE-constructed profile and a self-reported profile, split into different subreddits (top figure) and different similarity scores (bottom figure).}
    \label{fig:bf10-decomp-human}
\end{figure}

We create overviews of the different tests performed in Sections~\ref{sec:value-conflict-hypothesis} and~\ref{sec:value-conflict-hypothesis-human}. We decompose the aggregated scores into three separate figures, each showing how a single variable (either subreddit, similarity score, or profile threshold) impacts the obtained results. We show the decomposition for the $BF_{10}$ scores obtained for comparisons between two VPE-estimated profiles in Figures~\ref{fig:bf10-decomp} and for the comparison between VPE and self-reports in Figure~\ref{fig:bf10-decomp-human}. In the latter case, since we picked samples from Debagreement with authors with populated value profiles, we do not need to test over multiple profile thresholds.

We show the highest and lowest $BF_{10}$ scores and the test parameters in Tables~\ref{app:tab:top-5-vpe-vpe} and~\ref{app:tab:last-5-vpe-vpe} between two VPE profiles, and in Tables~\ref{app:tab:top-5-vpe-human} and~\ref{app:tab:last-5-vpe-human} for the experiments comparing VPE and self-reported profiles.

\begin{table}[!htb]
    \centering
    {\color{black}
    \begin{tabular}{@{}p{0.7cm}P{2cm}P{1.6cm}P{1.6cm}@{}}
    \toprule
         \textbf{$BF_{10}$} & \textbf{Subreddit} & \textbf{Similarity score} & \textbf{Profile threshold}  \\
    \midrule
        17.451 & BLM & CO & 10 \\
        12.485 & BLM & WC & 10 \\
        10.504 & BLM & $\tau$ & 250 \\
        4.223 & BLM & MD & 10 \\
        3.442 & Brexit & WC & 500\\
    \bottomrule
    \end{tabular}
    }
    \caption{The five tests between two VPE-constructed profiles with the highest $BF_{10}$ scores.}
    \label{app:tab:top-5-vpe-vpe}
\end{table}

\begin{table}[!htb]
    \centering
    {\color{black}
    \begin{tabular}{@{}p{0.7cm}P{2cm}P{1.6cm}P{1.6cm}@{}}
    \toprule
         \textbf{$BF_{10}$} & \textbf{Subreddit} & \textbf{Similarity score} & \textbf{Profile threshold}  \\
    \midrule
        0.079 & Brexit & MD & 50 \\
        0.081 & Brexit & $\tau$ & 50 \\
        0.083 & Brexit & $\tau$ & 10 \\
        0.085 & Brexit & $\tau$ & 1 \\
        0.086 & Brexit & MD & 10\\
    \bottomrule
    \end{tabular}
    }
    \caption{The five tests between two VPE-constructed profiles with the lowest $BF_{10}$ scores.}
    \label{app:tab:last-5-vpe-vpe}
\end{table}

\begin{table}[!htb]
    \centering
    {\color{black}
    \begin{tabular}{@{}p{0.7cm}P{2cm}P{1.6cm}P{1.6cm}@{}}
    \toprule
         \textbf{$BF_{10}$} & \textbf{Subreddit} & \textbf{Similarity score} \\
    \midrule
        6.490 & democrats & $\tau$ \\
        3.066 & democrats & MD \\
        2.543 & Brexit & MD \\
        2.407 & Brexit & CO \\
        2.230 & climate & CO\\
    \bottomrule
    \end{tabular}
    }
    \caption{The five tests between a VPE-constructed profile and a self-reported profile with the highest $BF_{10}$ scores.}
    \label{app:tab:top-5-vpe-human}
\end{table}

\begin{table}[!htb]
    \centering
    {\color{black}
    \begin{tabular}{@{}p{0.7cm}P{2cm}P{1.6cm}P{1.6cm}@{}}
    \toprule
         \textbf{$BF_{10}$} & \textbf{Subreddit} & \textbf{Similarity score} \\
    \midrule
        0.087 & republican & $\tau$ \\
        0.108 & Brexit & MD \\
        0.247 & Brexit & CO \\
        0.273 & Brexit & WC \\
        0.359 & repulican & MD\\
    \bottomrule
    \end{tabular}
    }
    \caption{The five tests between a VPE-constructed profile and a self-reported profile with the highest $BF_{10}$ scores.}
    \label{app:tab:last-5-vpe-human}
\end{table}

\subsection{Kendall \texorpdfstring{$\tau$}{t} vs. Spearman \texorpdfstring{$\rho$}{r}}
\begin{figure}[!htb]
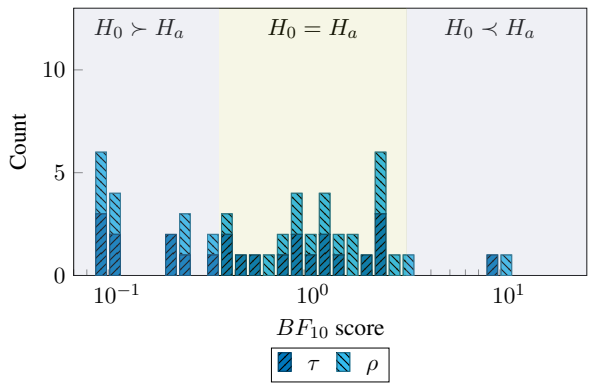

    \centering
    \includestandalone[width=\columnwidth]{tikz/spear_vs_tau}
    \caption{$BF_{10}$ scores when testing between two VPE-constructed profiles, obtained for the similarity scores Kendall $\tau$ and Spearman $\rho$.}
    \label{fig:bf10-spear-vs-tau}
\end{figure}

\mm{We include a comparative overview of the tests that use the Kendall $\tau$ and add the $BF_{10}$ scores for the same tests conducted with Spearman $\rho$. See Figure~\ref{fig:bf10-spear-vs-tau}. We see that generally, the $\rho$ scores are similarly distributed as the $\tau$ scores. Two tests that for $\tau$ fall into the undecidable range, for $\rho$ favor the null hypothesis $H_0$. We attribute this to the size of our value profiles: since we have only 10 entries, ties are likely, and Spearman $\rho$ does not explicitly account for them.}

\subsection{Agreement Analysis}
For additional results (\textbf{P}recision, \textbf{R}ecall, $F_1$ scores, accuracy, and the change w.r.t. a text-only baseline), see Table~\ref{tab:result-debagreement}.
\begin{table*}[!htb]
    \centering
    \begin{tabular}{@{}lccccc@{}}
        \toprule
         \textbf{Model} & \textbf{P} & \textbf{R} & \textbf{F1} & \textbf{Acc.} & $\Delta$ \textbf{F1}\\
        \midrule
        Majority                           & 0.12 & 0.33 & 0.18 & 0.37 & \\
        Only context ($\epsilon$)          & 0.21$_{\pm0.10}$ & 0.34$_{\pm0.01}$ & 0.24$_{\pm0.07}$ & 0.36$_{\pm0.00}$ & \\
        Only context ($z$)                 & 0.42$_{\pm0.00}$ & 0.41$_{\pm0.00}$ & 0.41$_{\pm0.00}$ & 0.43$_{\pm0.00}$ & \\
        Only context ($u$)                 & 0.33$_{\pm0.01}$ & 0.35$_{\pm0.00}$ & 0.31$_{\pm0.00}$ & 0.38$_{\pm0.00}$ & \\
        Only context ($v$)                 & 0.27$_{\pm0.00}$ & 0.37$_{\pm0.00}$ & 0.31$_{\pm0.00}$ & 0.40$_{\pm0.00}$ & \\
        TF-IDF + Logistic Regression       & 0.48$_{\pm0.01}$ & 0.47$_{\pm0.02}$ & 0.46$_{\pm0.03}$ & 0.48$_{\pm0.01}$ & --\\
        \hspace{1cm} + $\epsilon$          & 0.38$_{\pm0.01}$ & 0.37$_{\pm0.01}$ & 0.33$_{\pm0.05}$ & 0.36$_{\pm0.03}$ & -0.12 \\
        \hspace{1cm} + $z$                 & 0.51$_{\pm0.02}$ & 0.47$_{\pm0.04}$ & 0.43$_{\pm0.09}$ & 0.45$_{\pm0.06}$ & -0.03 \\
        \hspace{1cm} + $u$                 & 0.37$_{\pm0.00}$ & 0.36$_{\pm0.00}$ & 0.36$_{\pm0.01}$ & 0.36$_{\pm0.01}$ & -0.12 \\
        \hspace{1cm} + $v$                 & 0.51$_{\pm0.01}$ & 0.45$_{\pm0.02}$ & 0.41$_{\pm0.05}$ & 0.45$_{\pm0.04}$ & -0.04 \\
        BERT(-base-uncased)                & 0.62$_{\pm0.00}$ & 0.62$_{\pm0.01}$ & 0.62$_{\pm0.01}$ & 0.63$_{\pm0.01}$ & -- \\
        \hspace{1cm} + $\epsilon$          & 0.63$_{\pm0.00}$ & 0.62$_{\pm0.00}$ & 0.62$_{\pm0.00}$ & 0.64$_{\pm0.00}$ & 0.00 \\
        \hspace{1cm} + $z$                 & 0.63$_{\pm0.00}$ & 0.63$_{\pm0.00}$ & 0.63$_{\pm0.00}$ & 0.63$_{\pm0.00}$ & 0.01 \\
        \hspace{1cm} + $u$                 & 0.62$_{\pm0.00}$ & 0.62$_{\pm0.01}$ & 0.62$_{\pm0.01}$ & 0.63$_{\pm0.00}$ & 0.00   \\
        \hspace{1cm} + $v$                 & 0.64$_{\pm0.01}$ & 0.64$_{\pm0.01}$ & 0.64$_{\pm0.01}$ & 0.65$_{\pm0.01}$ & 0.02 \\
        \bottomrule
    \end{tabular}
    \caption{Performance of the agreement classification on a subset of Debagreement (sentence pairs for which both users were available on Reddit).}
    \label{tab:result-debagreement}
\end{table*}

\end{document}

%% file: commenting_macros.tex
\usepackage{xcolor}
\usepackage{amssymb}

\definecolor{pmcolor}{rgb}{0.4,0.2,0.1}
\definecolor{mmcolor}{rgb}{0.1,0.2,0.4}
\definecolor{tkcolor}{rgb}{0.2,0.4,0.1}
\definecolor{kncolor}{rgb}{0.3,0.1,0.3}

\definecolor{todocolor}{rgb}{0.9,0.1,0.1}
\definecolor{changedcolor}{rgb}{0.42,0.27,0.57}
\definecolor{removedcolor}{rgb}{0.867,0.176,0.361}

\newcommand{\mm}[1]{\textcolor{black}{#1}}



%% file: tikz/mds.tex
\definecolor{bblue}{HTML}{5cb1eb}
\definecolor{rred}{HTML}{be1908}
\definecolor{ggreen}{HTML}{aaad00}
\definecolor{dblue}{HTML}{001158}

\begin{tikzpicture}
\begin{axis}[
    enlargelimits=0.2,
    ticks=none,
    axis x line=bottom,
    axis y line=left,
    xlabel={Dimension 1}, 
    ylabel={Dimension 2},
    xmin=-0.35,
    xmax=0.25,
    ymin=-0.27,
    ymax=0.25,  
    yticklabel style={
            /pgf/number format/fixed,
            /pgf/number format/precision=5
    },
    scaled y ticks=false,
    x tick label style={
        /pgf/number format/.cd,
            fixed,
            fixed zerofill,
            precision=2,
        /tikz/.cd
    },
    typeset ticklabels with strut,
    legend columns=2, 
    legend style={at={(1.05,-0.13)},anchor=north east},
    legend cell align=left,
    smooth,
]
    \addplot[
        scatter/classes={self-enhancement={bblue}, self-transcendence={rred}, openness-to-change={ggreen}, conservation={dblue}},
        mark size=5pt,
        scatter, mark=*, only marks, 
        scatter src=explicit symbolic,
        nodes near coords*={\Label},
        visualization depends on={value \thisrow{label} \as \Label} 
    ] table [meta=class] {
    x y label class
    -0.038845 -0.241918 benevolence self-transcendence
    -0.114034 0.043709 conformity conservation
    0.075738 0.125663 security conservation
    -0.063512 0.188132 stimulation openness-to-change
    -0.286490 -0.009333 tradition conservation
    0.237122 -0.158605 universalism self-transcendence
    0.021125 0.015197 power self-enhancement
    
    };
    
    \addplot[
        scatter/classes={self-enhancement={bblue}, self-transcendence={rred}, openness-to-change={ggreen}, conservation={dblue}},
        mark size=5pt,
        scatter, mark=*, only marks, 
        scatter src=explicit symbolic,
    ] table [meta=class] {
    x y label class
    0.003716 -0.004469 achievement self-enhancement
    -0.011487 -0.048177 self-direction openness-to-change

    };
    
    \addplot[
        mark size=5pt,
        draw=bblue,
        fill=ggreen,
        mark=halfcircle*, 
        mark color=bblue,
        mark options={rotate=-90},
        nodes near coords={\Label},
        visualization depends on={value \thisrow{label} \as \Label} 
    ] table [] {
    x y label
    0.176666 0.089800 hedonism self
    };
    
    \node[] at (0.0037, -0.024) {achievement};
    \node[] at (-0.015, -0.075 ) {self-direction};
 
    \legend{
        self-enhancement,
        self-transcendence,
        openness-to-change,
        conservation,
    }
\end{axis}
\end{tikzpicture}

%% file: tikz/bayesfactors.tex
\pgfplotstableread{
x y
0.08678646731304027 6
0.10389431518512199 11
0.12437456048131589 3
0.14889198958920244 9
0.17824243541476853 4
0.2133786100262569 8
0.25544102957743214 5
0.30579503532968616 7
0.36607511247106855 5
0.4382379453159678 3
0.5246259310510748 8
0.6280432136764552 5
0.751846706194645 3
0.9000550556174705 3
1.0774790877820806 1
1.2898779661997932 4
1.5441461338358837 1
1.8485371059288118 1
2.212931378137845 3
2.6491571462864907 2
3.1713742482274245 4
3.796533790537312 0
4.5449283794707735 1
5.44085081660643 0
6.51338264036933 0
7.797338109396161 0
9.334394269333796 0
11.174443784908433 1
13.377214450037444 1
16.01420794509433 1

}\loadedtable

\begin{tikzpicture}
    \begin{axis}[
            ybar=3pt,
            bar width=5pt,
            xmode=log,
            log x ticks with fixed point/.style={
                xticklabel={
                    \pgfkeys{/pgf/fpu=true}
                    \pgfmathparse{exp(\tick)}%
                    \pgfmathprintnumber[fixed relative, precision=3]{\pgfmathresult}
                    \pgfkeys{/pgf/fpu=false}
                }
            },
            width=10cm,
            height=6.6cm,
            ymin=0,
            ymax=13,
            xmin=0.06,
            xlabel=$BF_{10}$ score,
            ylabel=Count,
            xmax=25,
            xtick pos=left,
            ytick pos=left,
        ]
        \begin{scope}[on background layer]
            \fill[dblue!30, opacity=0.2] (0.06,0) rectangle (0.33,17);
            \fill[ggreen, opacity=0.1] (0.33,0) rectangle (3,17);
            \fill[dblue!30, opacity=0.2] (3,0) rectangle (33,17);
        \end{scope}
        \addplot[draw=bblue!50!blue,fill=bblue!50] table[x=x, y=y] {\loadedtable};    
        \node[] at (axis cs: 1,12) {$H_0 = H_a$};
        \node[] at (axis cs: 0.13,12) {$H_0 \succ H_a$};
        \node[] at (axis cs: 8,12) {$H_0 \prec H_a$};

    \end{axis}
\end{tikzpicture}

%% file: tikz/bayesfactors_human.tex
\pgfplotstableread{
x y
0.0937242474892384 1
0.1082121793480952 1
0.12493966153858821 0
0.1442528846513962 0
0.16655155355789159 0
0.19229715966916533 0
0.2220225319241783 0
0.2563428641735004 2
0.2959684471804623 0
0.3417193687401638 1
0.3945424861487886 0
0.45553102228405007 2
0.52594719085563 0
0.6072483190759281 6
0.701117008383757 4
0.8094959574248334 14
0.9346281679825937 9
1.0791033659579605 2
1.2459115981228233 18
1.4385051138812233 11
1.660869812738061 9
1.9176077361462398 1
2.2140323109767106 3
2.556278315762409 2
2.9514288455683855 1
3.407661903142616 0
3.9344196501858377 0
4.54260382155072 0
5.244801346646518 0
6.055544847491164 1

}\loadedtable

\begin{tikzpicture}
    \begin{axis}[
            ybar=0pt,
            bar width=7pt,
            xmode=log,
            log x ticks with fixed point/.style={
                xticklabel={
                    \pgfkeys{/pgf/fpu=true}
                    \pgfmathparse{exp(\tick)}%
                    \pgfmathprintnumber[fixed relative, precision=3]{\pgfmathresult}
                    \pgfkeys{/pgf/fpu=false}
                }
            },
            width=10cm,
            height=6.6cm,
            ymin=0,
            xmin=0.06,
            ymax=20,
            xmax=10,
            xlabel=$BF_{10}$ score,
            ylabel=Count,
            xtick pos=left,
            ytick pos=left,
        ]
        \begin{scope}[on background layer]
            \fill[dblue!30, opacity=0.2] (0.06,0) rectangle (0.33,  20);
            \fill[ggreen, opacity=0.1] (0.33,0) rectangle (3,       20);
            \fill[dblue!30, opacity=0.2] (3,0) rectangle (10,       20);
        \end{scope}
        \addplot[draw=bblue!50!blue,fill=bblue!50] table[x=x, y=y] {\loadedtable};    
        \node[] at (axis cs: 1,    18.8) {$H_0 = H_a$};
        \node[] at (axis cs: 0.13, 18.8) {$H_0 \succ H_a$};
        \node[] at (axis cs: 5,    18.8) {$H_0 \prec H_a$};

    \end{axis}
\end{tikzpicture}

%% file: tikz/use_case_agreement.tex
\pgfplotsset{
    /pgfplots/my legend/.style={
        legend image code/.code={
                \draw[black, dashed](-0.05cm,0cm) -- (0.3cm,0cm);%
           }
      }
}
\definecolor{bblue}{HTML}{5cb1eb}
\definecolor{rred}{HTML}{be1908}
\definecolor{ggreen}{HTML}{aaad00}

\begin{tikzpicture}
    \begin{axis}[
            ybar=3pt,
            ymajorgrids = true,
            xtick = data,
            enlarge x limits=0.15,
            symbolic x coords={text only,$\epsilon$,$z$,$u$,$v$},
            y tick label style={
                /pgf/number format/.cd,
                fixed,
                fixed zerofill,
                precision=2,
            },
            width=10cm,
            height=6.6cm,
            ymin=0,
            ymax=0.8,
            xlabel=Context,
            ylabel=F1 score,
            ytick pos=left,
            legend cell align=left,
            legend style={
                    at={(1,1.05)},
                    legend columns=4,
                    anchor=south east,
                    column sep=1ex
            },
            point meta=explicit symbolic,
            nodes near coords
        ]

        \addplot[style={bblue!50!blue,fill=bblue!30,mark=none}]
            coordinates {($\epsilon$, 0.24) ($z$, 0.41) ($u$, 0.31) ($v$, 0.31) (text only, 0)};

        \addplot[style={rred!70!black,fill=rred!30,mark=none}]
            coordinates {($\epsilon$, 0.33)[-{}-] ($z$,0.43)[-] ($u$,0.36)[-{}-]  ($v$,0.41)[-] (text only, 0.46)};

        \addplot[style={ggreen!70!black,fill=ggreen!30,mark=none}]
            coordinates {($\epsilon$, 0.62)[=] ($z$, 0.63)[+] ($u$, 0.62)[=] ($v$, 0.64)[+]  (text only, 0.62)};

        \legend{Context only,TF-IDF,BERT}
        \addlegendimage{my legend}
        \addlegendentry{Majority baseline}

        \coordinate (A) at (axis cs:$\epsilon$,0.18);
        \coordinate (O1) at (rel axis cs:0,0);
        \coordinate (O2) at (rel axis cs:1,0);
        \draw [black,sharp plot,dashed] (A -| O1) -- (A -| O2);

    \end{axis}
\end{tikzpicture}

%% file: tikz/BF_colored.tex
\pgfplotstableread{
x republican democrats climate brexit blacklivesmatter
0.08678646731304027 0 0 0 6 0
0.10389431518512199 0 9 0 2 0
0.12437456048131589 0 2 0 1 0
0.14889198958920244 6 0 3 0 0
0.17824243541476853 1 2 0 1 0
0.2133786100262569 2 3 2 1 0
0.25544102957743214 3 1 1 0 0
0.30579503532968616 1 1 1 1 3
0.36607511247106855 0 1 2 0 2
0.4382379453159678 0 0 2 1 0
0.5246259310510748 0 0 2 2 4
0.6280432136764552 2 1 0 1 1
0.751846706194645 3 0 0 0 0
0.9000550556174705 1 0 2 0 0
1.0774790877820806 0 0 0 1 0
1.2898779661997932 1 0 2 1 0
1.5441461338358837 0 0 0 0 1
1.8485371059288118 0 0 0 0 1
2.212931378137845 0 0 1 0 2
2.6491571462864907 0 0 1 1 0
3.1713742482274245 0 0 1 1 2
3.796533790537312 0 0 0 0 0
4.5449283794707735 0 0 0 0 1
5.44085081660643 0 0 0 0 0
6.51338264036933 0 0 0 0 0
7.797338109396161 0 0 0 0 0
9.334394269333796 0 0 0 0 0
11.174443784908433 0 0 0 0 1
13.377214450037444 0 0 0 0 1
16.01420794509433 0 0 0 0 1
}\loadedtable

\pgfplotstableread{
x schwartz_soft_cosine kendall cosine absolute_error
0.08678646731304027 0 3 0 3
0.10389431518512199 3 2 3 3
0.12437456048131589 1 0 1 1
0.14889198958920244 2 0 2 5
0.17824243541476853 2 0 1 1
0.2133786100262569 1 1 2 4
0.25544102957743214 1 1 3 0
0.30579503532968616 2 1 1 3
0.36607511247106855 2 1 1 1
0.4382379453159678 1 0 2 0
0.5246259310510748 4 0 4 0
0.6280432136764552 1 1 0 3
0.751846706194645 0 3 0 0
0.9000550556174705 1 1 1 0
1.0774790877820806 0 0 1 0
1.2898779661997932 1 3 0 0
1.5441461338358837 0 1 0 0
1.8485371059288118 0 1 0 0
2.212931378137845 0 3 0 0
2.6491571462864907 0 1 1 0
3.1713742482274245 2 1 1 0
3.796533790537312 0 0 0 0
4.5449283794707735 0 0 0 1
5.44085081660643 0 0 0 0
6.51338264036933 0 0 0 0
7.797338109396161 0 0 0 0
9.334394269333796 0 0 0 0
11.174443784908433 0 1 0 0
13.377214450037444 1 0 0 0
16.01420794509433 0 0 1 0
}\loadedtablesm

\pgfplotstableread{
x ff fh twof one ten
0.08678646731304027 2 0 0 2 2
0.10389431518512199 1 3 3 3 1
0.12437456048131589 0 0 1 0 2
0.14889198958920244 0 4 4 0 1
0.17824243541476853 2 1 0 1 0
0.2133786100262569 2 1 1 3 1
0.25544102957743214 2 0 1 0 2
0.30579503532968616 3 0 2 2 0
0.36607511247106855 1 1 0 1 2
0.4382379453159678 0 2 0 0 1
0.5246259310510748 0 2 2 3 1
0.6280432136764552 0 1 1 2 1
0.751846706194645 0 0 1 1 1
0.9000550556174705 2 1 0 0 0
1.0774790877820806 0 0 1 0 0
1.2898779661997932 1 1 2 0 0
1.5441461338358837 1 0 0 0 0
1.8485371059288118 0 0 0 1 0
2.212931378137845 1 1 0 0 1
2.6491571462864907 0 1 0 0 1
3.1713742482274245 2 1 0 1 0
3.796533790537312 0 0 0 0 0
4.5449283794707735 0 0 0 0 1
5.44085081660643 0 0 0 0 0
6.51338264036933 0 0 0 0 0
7.797338109396161 0 0 0 0 0
9.334394269333796 0 0 0 0 0
11.174443784908433 0 0 1 0 0
13.377214450037444 0 0 0 0 1
16.01420794509433 0 0 0 0 1
}\loadedtablet

\begin{tikzpicture}
    \begin{axis}[
            ybar stacked,
            name=bf-subreddits,    
            anchor=east,
            bar width=5pt,
            xmode=log,
            log x ticks with fixed point/.style={
                xticklabel={
                    \pgfkeys{/pgf/fpu=true}
                    \pgfmathparse{exp(\tick)}%
                    \pgfmathprintnumber[fixed relative, precision=3]{\pgfmathresult}
                    \pgfkeys{/pgf/fpu=false}
                }
            },
            width=10cm,
            height=6cm,
            ymin=0,
            ymax=13,
            xmin=0.06,
            xlabel=$BF_{10}$ score,
            ylabel=Count,
            xmax=25,
            xtick pos=left,
            ytick pos=left,
            legend cell align=left,
            legend style={
                    at={(0.5,-0.4)},
                    legend columns=5,
                    anchor=south,
                    column sep=0.5em
            },
        ]
        \fill[dblue!30, opacity=0.1] (0.06,0) rectangle (0.33,13);
        \fill[ggreen, opacity=0.1] (0.33,0) rectangle (3,13);
        \fill[dblue!30, opacity=0.1] (3,0) rectangle (33,13);
        \addplot[draw=ibm1!60!black,fill=ibm1, postaction={pattern=north east lines}] table[x=x, y=democrats] {\loadedtable};    
        \addplot[draw=ibm2!60!black,fill=ibm2, postaction={pattern=north west lines}] table[x=x, y=brexit] {\loadedtable};    
        \addplot[draw=ibm3!60!black,fill=ibm3, postaction={pattern=horizontal lines}] table[x=x, y=blacklivesmatter] {\loadedtable};    
        \addplot[draw=ibm4!60!black,fill=ibm4, postaction={pattern=vertical lines}  ] table[x=x, y=climate] {\loadedtable};    
        \addplot[draw=ibm5!60!black,fill=ibm5, postaction={pattern=north east lines}] table[x=x, y=republican] {\loadedtable};    
        \node[] at (axis cs: 1,12) {$H_0 = H_a$};
        \node[] at (axis cs: 0.13,12) {$H_0 \succ H_a$};
        \node[] at (axis cs: 8,12) {$H_0 \prec H_a$};
        \legend{democrats, brexit, BLM, climate, republican};
    \end{axis}

    \begin{axis}[
            ybar stacked,
            name=bf-sm,    
            anchor=east,
    	    at={($(bf-subreddits.east)-(0cm,7cm)$)},
            bar width=5pt,
            xmode=log,
            log x ticks with fixed point/.style={
                xticklabel={
                    \pgfkeys{/pgf/fpu=true}
                    \pgfmathparse{exp(\tick)}%
                    \pgfmathprintnumber[fixed relative, precision=3]{\pgfmathresult}
                    \pgfkeys{/pgf/fpu=false}
                }
            },
            width=10cm,
            height=6cm,
            ymin=0,
            ymax=13,
            xmin=0.06,
            xlabel=$BF_{10}$ score,
            ylabel=Count,
            xmax=25,
            xtick pos=left,
            ytick pos=left,
            legend cell align=left,
            legend style={
                    at={(0.5,-0.4)},
                    legend columns=4,
                    anchor=south,
                    column sep=0.5em
            },
        ]
        \fill[dblue!30, opacity=0.2] (0.06,0) rectangle (0.33,13);
        \fill[ggreen, opacity=0.1] (0.33,0) rectangle (3,13);
        \fill[dblue!30, opacity=0.2] (3,0) rectangle (33,13);
        \addplot[draw=ibm1!60!black,fill=ibm1, postaction={pattern=north east lines}] table[x=x, y=kendall] {\loadedtablesm};    
        \addplot[draw=ibm2!60!black,fill=ibm2, postaction={pattern=north west lines}] table[x=x, y=absolute_error] {\loadedtablesm};    
        \addplot[draw=ibm3!60!black,fill=ibm3, postaction={pattern=horizontal lines}] table[x=x, y=schwartz_soft_cosine] {\loadedtablesm};
        \addplot[draw=ibm4!60!black,fill=ibm4, postaction={pattern=vertical lines}  ] table[x=x, y=cosine] {\loadedtablesm};    
        \node[] at (axis cs: 1,12) {$H_0 = H_a$};
        \node[] at (axis cs: 0.13,12) {$H_0 \succ H_a$};
        \node[] at (axis cs: 8,12) {$H_0 \prec H_a$};
        \legend{$\tau$, $AE$, $WC$, $CS$};
    \end{axis}

        \begin{axis}[
            ybar stacked,
            name=bf-t,    
            anchor=east,
    	    at={($(bf-sm.east)-(0cm,7cm)$)},
            bar width=5pt,
            xmode=log,
            log x ticks with fixed point/.style={
                xticklabel={
                    \pgfkeys{/pgf/fpu=true}
                    \pgfmathparse{exp(\tick)}%
                    \pgfmathprintnumber[fixed relative, precision=3]{\pgfmathresult}
                    \pgfkeys{/pgf/fpu=false}
                }
            },
            width=10cm,
            height=6cm,
            ymin=0,
            ymax=13,
            xmin=0.06,
            xlabel=$BF_{10}$ score,
            ylabel=Count,
            xmax=25,
            xtick pos=left,
            ytick pos=left,
            legend cell align=left,
            legend style={
                    at={(0.5,-0.4)},
                    legend columns=5,
                    anchor=south,
                    column sep=0.5em
            },
        ]
        \fill[dblue!30, opacity=0.2] (0.06,0) rectangle (0.33,13);
        \fill[ggreen, opacity=0.1] (0.33,0) rectangle (3,13);
        \fill[dblue!30, opacity=0.2] (3,0) rectangle (33,13);
        \addplot[draw=seq5!60!black,fill=seq5, postaction={pattern=north east lines}] table[x=x, y=fh] {\loadedtablet};    
        \addplot[draw=seq4!60!black,fill=seq4, postaction={pattern=north west lines}] table[x=x, y=twof] {\loadedtablet};    
        \addplot[draw=seq3!60!black,fill=seq3, postaction={pattern=horizontal lines}] table[x=x, y=ff] {\loadedtablet};    
        \addplot[draw=seq2!60!black,fill=seq2, postaction={pattern=vertical lines}  ] table[x=x, y=ten] {\loadedtablet};    
        \addplot[draw=seq1!60!black,fill=seq1, postaction={pattern=north east lines}] table[x=x, y=one] {\loadedtablet};    
        \node[] at (axis cs: 1,12) {$H_0 = H_a$};
        \node[] at (axis cs: 0.13,12) {$H_0 \succ H_a$};
        \node[] at (axis cs: 8,12) {$H_0 \prec H_a$};
        \legend{500,250,50,10,1};
    \end{axis}
\end{tikzpicture}

%% file: tikz/BF_human_colored.tex
\pgfplotstableread{
x republican democrats climate brexit blacklivesmatter
0.0937242474892384 1 0 0 0 0
0.1082121793480952 0 0 0 1 0
0.12493966153858821 0 0 0 0 0
0.1442528846513962 0 0 0 0 0
0.16655155355789159 0 0 0 0 0
0.19229715966916533 0 0 0 0 0
0.2220225319241783 0 0 0 0 0
0.2563428641735004 0 0 0 2 0
0.2959684471804623 0 0 0 0 0
0.3417193687401638 1 0 0 0 0
0.3945424861487886 0 0 0 0 0
0.45553102228405007 2 0 0 0 0
0.52594719085563 0 0 0 0 0
0.6072483190759281 0 2 1 2 1
0.701117008383757 1 0 0 1 2
0.8094959574248334 1 2 4 2 5
0.9346281679825937 0 3 0 2 4
1.0791033659579605 0 1 0 0 1
1.2459115981228233 5 5 4 0 4
1.4385051138812233 4 1 0 1 5
1.660869812738061 5 0 1 2 1
1.9176077361462398 0 0 0 0 1
2.2140323109767106 0 0 2 1 0
2.556278315762409 0 0 0 2 0
2.9514288455683855 0 1 0 0 0
3.407661903142616 0 0 0 0 0
3.9344196501858377 0 0 0 0 0
4.54260382155072 0 0 0 0 0
5.244801346646518 0 0 0 0 0
6.055544847491164 0 1 0 0 0
}\loadedtable

\pgfplotstableread{
x schwartz_soft_cosine cosine absolute_error kendall
0.0937242474892384 0 0 0 1
0.1082121793480952 0 0 1 0
0.12493966153858821 0 0 0 0
0.1442528846513962 0 0 0 0
0.16655155355789159 0 0 0 0
0.19229715966916533 0 0 0 0
0.2220225319241783 0 0 0 0
0.2563428641735004 1 1 0 0
0.2959684471804623 0 0 0 0
0.3417193687401638 0 0 1 0
0.3945424861487886 0 0 0 0
0.45553102228405007 1 1 0 0
0.52594719085563 0 0 0 0
0.6072483190759281 0 0 3 3
0.701117008383757 1 0 1 2
0.8094959574248334 3 4 3 4
0.9346281679825937 3 3 2 1
1.0791033659579605 0 0 0 2
1.2459115981228233 6 5 4 3
1.4385051138812233 2 4 3 2
1.660869812738061 3 2 1 3
1.9176077361462398 0 0 1 0
2.2140323109767106 2 1 0 0
2.556278315762409 0 1 1 0
2.9514288455683855 0 0 1 0
3.407661903142616 0 0 0 0
3.9344196501858377 0 0 0 0
4.54260382155072 0 0 0 0
5.244801346646518 0 0 0 0
6.055544847491164 0 0 0 1
}\loadedtablesm

\begin{tikzpicture}

    \begin{axis}[
            ybar stacked,
            name=bf-subreddits,
            anchor=east,
            bar width=7pt,
            xmode=log,
            log x ticks with fixed point/.style={
                xticklabel={
                    \pgfkeys{/pgf/fpu=true}
                    \pgfmathparse{exp(\tick)}%
                    \pgfmathprintnumber[fixed relative, precision=3]{\pgfmathresult}
                    \pgfkeys{/pgf/fpu=false}
                }
            },
            width=10cm,
            height=6.6cm,
            ymin=0,
            ymax=20,
            xmin=0.06,
            xlabel=$BF_{10}$ score,
            ylabel=Count,
            xmax=10,
            xtick pos=left,
            ytick pos=left,
            legend cell align=left,
            legend style={
                    at={(0.5,-0.4)},
                    legend columns=5,
                    anchor=south,
                    column sep=0.5em
            },
        ]
        \fill[dblue!30, opacity=0.2] (0.06,0) rectangle (0.33,20);
        \fill[ggreen, opacity=0.1] (0.33,0) rectangle (3,20);
        \fill[dblue!30, opacity=0.2] (3,0) rectangle (33,20);
        \addplot[draw=ibm1!60!black,fill=ibm1, postaction={pattern=north east lines}] table[x=x, y=republican] {\loadedtable};
        \addplot[draw=ibm2!60!black,fill=ibm2, postaction={pattern=north west lines}] table[x=x, y=climate] {\loadedtable};
        \addplot[draw=ibm3!60!black,fill=ibm3, postaction={pattern=horizontal lines}] table[x=x, y=blacklivesmatter] {\loadedtable};
        \addplot[draw=ibm4!60!black,fill=ibm4, postaction={pattern=vertical lines}  ] table[x=x, y=brexit] {\loadedtable};
        \addplot[draw=ibm5!60!black,fill=ibm5, postaction={pattern=north east lines}] table[x=x, y=democrats] {\loadedtable};
        \node[] at (axis cs: 1,    18.8) {$H_0 = H_a$};
        \node[] at (axis cs: 0.13, 18.8) {$H_0 \succ H_a$};
        \node[] at (axis cs: 5,    18.8) {$H_0 \prec H_a$};
        \legend{republican, climate, BLM, brexit, democrats};
    \end{axis}
    \begin{axis}[
            ybar stacked,
            name=bf-sm,
            anchor=east,
            at={($(bf-subreddits.east)-(0cm,8cm)$)},
            bar width=7pt,
            xmode=log,
            log x ticks with fixed point/.style={
                xticklabel={
                    \pgfkeys{/pgf/fpu=true}
                    \pgfmathparse{exp(\tick)}%
                    \pgfmathprintnumber[fixed relative, precision=3]{\pgfmathresult}
                    \pgfkeys{/pgf/fpu=false}
                }
            },
            width=10cm,
            height=6.6cm,
            ymin=0,
            ymax=20,
            xmin=0.06,
            xlabel=$BF_{10}$ score,
            ylabel=Count,
            xmax=10,
            xtick pos=left,
            ytick pos=left,
            legend cell align=left,
            legend style={
                    at={(0.5,-0.4)},
                    legend columns=4,
                    anchor=south,
                    column sep=0.5em
            },
        ]
        \fill[dblue!30, opacity=0.2] (0.06,0) rectangle (0.33,20);
        \fill[ggreen, opacity=0.1] (0.33,0) rectangle (3,20);
        \fill[dblue!30, opacity=0.2] (3,0) rectangle (33,20);
        \addplot[draw=ibm1!60!black,fill=ibm1, postaction={pattern=north east lines}] table[x=x, y=kendall] {\loadedtablesm};
        \addplot[draw=ibm2!60!black,fill=ibm2, postaction={pattern=north west lines}] table[x=x, y=absolute_error] {\loadedtablesm};
        \addplot[draw=ibm3!60!black,fill=ibm3, postaction={pattern=horizontal lines}] table[x=x, y=schwartz_soft_cosine] {\loadedtablesm};
        \addplot[draw=ibm4!60!black,fill=ibm4, postaction={pattern=vertical lines}  ] table[x=x, y=cosine] {\loadedtablesm};
        \node[] at (axis cs: 1,    18.8) {$H_0 = H_a$};
        \node[] at (axis cs: 0.13, 18.8) {$H_0 \succ H_a$};
        \node[] at (axis cs: 5,    18.8) {$H_0 \prec H_a$};
        \legend{$\tau$, $AE$, $WC$, $CS$};
    \end{axis}
\end{tikzpicture}

%% file: tikz/spear_vs_tau.tex
\pgfplotstableread{
x republican democrats climate brexit blacklivesmatter
0.08678646731304027 0 0 0 6 0
0.10389431518512199 0 9 0 2 0
0.12437456048131589 0 2 0 1 0
0.14889198958920244 6 0 3 0 0
0.17824243541476853 1 2 0 1 0
0.2133786100262569 2 3 2 1 0
0.25544102957743214 3 1 1 0 0
0.30579503532968616 1 1 1 1 3
0.36607511247106855 0 1 2 0 2
0.4382379453159678 0 0 2 1 0
0.5246259310510748 0 0 2 2 4
0.6280432136764552 2 1 0 1 1
0.751846706194645 3 0 0 0 0
0.9000550556174705 1 0 2 0 0
1.0774790877820806 0 0 0 1 0
1.2898779661997932 1 0 2 1 0
1.5441461338358837 0 0 0 0 1
1.8485371059288118 0 0 0 0 1
2.212931378137845 0 0 1 0 2
2.6491571462864907 0 0 1 1 0
3.1713742482274245 0 0 1 1 2
3.796533790537312 0 0 0 0 0
4.5449283794707735 0 0 0 0 1
5.44085081660643 0 0 0 0 0
6.51338264036933 0 0 0 0 0
7.797338109396161 0 0 0 0 0
9.334394269333796 0 0 0 0 0
11.174443784908433 0 0 0 0 1
13.377214450037444 0 0 0 0 1
16.01420794509433 0 0 0 0 1
}\loadedtable

\pgfplotstableread{
x spearman kendall
0.0827852463326178 3 3
0.09756730657138772 2 2
0.11498883838972734 0 0
0.13552114349435573 0 0
0.15971967880717805 0 0
0.18823908314594903 2 0
0.22185088705572237 1 2
0.2614643848921131 0 0
0.3081512338052547 1 1
0.36317444509653773 2 1
0.42802255224632163 1 0
0.5044498799544023 1 0
0.5945240035846681 0 1
0.700681682926145 1 1
0.825794783436858 2 2
0.9732479683265909 1 1
1.14703026326911 2 2
1.3518429708282775 1 1
1.5932268539884809 0 2
1.877712029463574 1 0
2.21299462582226 3 3
2.6081449855318226 0 1
3.0738530433743314 0 1
3.6227175194154286 0 0
4.269586750013422 0 0
5.031960377311374 0 0
5.930462764049008 0 0
6.989400940904007 0 0
8.23742218041623 1 0
9.708288986729224 0 1
}\loadedtablesm

\begin{tikzpicture}

    \begin{axis}[
            ybar stacked,
            name=bf-sm,    
            anchor=east,
            bar width=5pt,
            xmode=log,
            log x ticks with fixed point/.style={
                xticklabel={
                    \pgfkeys{/pgf/fpu=true}
                    \pgfmathparse{exp(\tick)}%
                    \pgfmathprintnumber[fixed relative, precision=3]{\pgfmathresult}
                    \pgfkeys{/pgf/fpu=false}
                }
            },
            width=10cm,
            height=6cm,
            ymin=0,
            ymax=13,
            xmin=0.06,
            xlabel=$BF_{10}$ score,
            ylabel=Count,
            xmax=25,
            xtick pos=left,
            ytick pos=left,
            legend cell align=left,
            legend style={
                    at={(0.5,-0.4)},
                    legend columns=4,
                    anchor=south,
                    column sep=0.5em
            },
        ]
        \fill[dblue!30, opacity=0.2] (0.06,0) rectangle (0.33,13);
        \fill[ggreen, opacity=0.1] (0.33,0) rectangle (3,13);
        \fill[dblue!30, opacity=0.2] (3,0) rectangle (33,13);
        \addplot[draw=ibm1!60!black,fill=ibm1, postaction={pattern=north east lines}] table[x=x, y=spearman] {\loadedtablesm};    
        \addplot[draw=ibm2!60!black,fill=ibm2, postaction={pattern=north west lines}] table[x=x, y=kendall] {\loadedtablesm};    
        \node[] at (axis cs: 1,12) {$H_0 = H_a$};
        \node[] at (axis cs: 0.13,12) {$H_0 \succ H_a$};
        \node[] at (axis cs: 8,12) {$H_0 \prec H_a$};
        \legend{$\tau$, $\rho$};
    \end{axis}

\end{tikzpicture}

%% file: main.bbl
\begin{thebibliography}{82}
\expandafter\ifx\csname natexlab\endcsname\relax\def\natexlab#1{#1}\fi

\bibitem[{Akata et~al.(2020)Akata, Balliet, de~Rijke, Dignum, Dignum, Eiben,
  Fokkens, Grossi, Hindriks, Hoos, Hung, Jonker, Monz, Neerincx, Oliehoek,
  Prakken, Schlobach, van~der Gaag, van Harmelen, van Hoof, van Riemsdijk, van
  Wynsberghe, Verbrugge, Verheij, Vossen, and Welling}]{akata2020research}
Zeynep Akata, Dan Balliet, Maarten de~Rijke, Frank Dignum, Virginia Dignum,
  Guszti Eiben, Antske Fokkens, Davide Grossi, Koen Hindriks, Holger Hoos,
  Hayley Hung, Catholijn Jonker, Christof Monz, Mark Neerincx, Frans Oliehoek,
  Henry Prakken, Stefan Schlobach, Linda van~der Gaag, Frank van Harmelen,
  Herke van Hoof, Birna van Riemsdijk, Aimee van Wynsberghe, Rineke Verbrugge,
  Bart Verheij, Piek Vossen, and Max Welling. 2020.
\newblock \href {https://doi.org/10.1109/MC.2020.2996587} {A research agenda
  for hybrid intelligence: Augmenting human intellect with collaborative,
  adaptive, responsible, and explainable artificial intelligence}.
\newblock \emph{Computer}, 53(8):18--28.

\bibitem[{Al~Khatib et~al.(2020)Al~Khatib, V{\"o}lske, Syed, Kolyada, and
  Stein}]{al2020exploiting}
Khalid Al~Khatib, Michael V{\"o}lske, Shahbaz Syed, Nikolay Kolyada, and Benno
  Stein. 2020.
\newblock \href {https://doi.org/10.18653/v1/2020.acl-main.632} {Exploiting
  personal characteristics of debaters for predicting persuasiveness}.
\newblock In \emph{Proceedings of the 58th Annual Meeting of the Association
  for Computational Linguistics}, pages 7067--7072.

\bibitem[{ALDayel and Magdy(2021)}]{aldayel2021stance}
Abeer ALDayel and Walid Magdy. 2021.
\newblock \href {https://doi.org/https://doi.org/10.1016/j.ipm.2021.102597}
  {Stance detection on social media: State of the art and trends}.
\newblock \emph{Information Processing \& Management}, 58(4):102597.

\bibitem[{Allaway and McKeown(2020)}]{allaway2020zero}
Emily Allaway and Kathleen McKeown. 2020.
\newblock \href {https://doi.org/10.18653/v1/2020.emnlp-main.717}
  {{Z}ero-{S}hot {S}tance {D}etection: {A} {D}ataset and {M}odel using
  {G}eneralized {T}opic {R}epresentations}.
\newblock In \emph{Proceedings of the 2020 Conference on Empirical Methods in
  Natural Language Processing (EMNLP)}, pages 8913--8931, Online. Association
  for Computational Linguistics.

\bibitem[{Alshomary et~al.(2022)Alshomary, El~Baff, Gurcke, and
  Wachsmuth}]{alshomary2022moral}
Milad Alshomary, Roxanne El~Baff, Timon Gurcke, and Henning Wachsmuth. 2022.
\newblock \href {https://doi.org/10.18653/v1/2022.acl-long.601} {The moral
  debater: A study on the computational generation of morally framed
  arguments}.
\newblock In \emph{Proceedings of the 60th Annual Meeting of the Association
  for Computational Linguistics (Volume 1: Long Papers)}, pages 8782--8797,
  Dublin, Ireland. Association for Computational Linguistics.

\bibitem[{Azer et~al.(2020)Azer, Khashabi, Sabharwal, and Roth}]{azer2020not}
Erfan~Sadeqi Azer, Daniel Khashabi, Ashish Sabharwal, and Dan Roth. 2020.
\newblock \href {https://aclanthology.org/2020.acl-main.506.pdf} {Not all
  claims are created equal: Choosing the right statistical approach to assess
  hypotheses}.
\newblock In \emph{Proceedings of the 58th Annual Meeting of the Association
  for Computational Linguistics}, pages 5715--5725.

\bibitem[{Bail et~al.(2018)Bail, Argyle, Brown, Bumpus, Chen, Hunzaker, Lee,
  Mann, Merhout, and Volfovsky}]{bail2018exposure}
Christopher~A. Bail, Lisa~P. Argyle, Taylor~W. Brown, John~P. Bumpus, Haohan
  Chen, M.~B.~Fallin Hunzaker, Jaemin Lee, Marcus Mann, Friedolin Merhout, and
  Alexander Volfovsky. 2018.
\newblock \href {https://doi.org/10.1073/pnas.1804840115} {Exposure to opposing
  views on social media can increase political polarization}.
\newblock \emph{Proceedings of the National Academy of Sciences},
  115(37):9216--9221.

\bibitem[{Beck et~al.(2019)Beck, Neupane, and Carroll}]{beck2019managing}
Jordan Beck, Bikalpa Neupane, and John~M. Carroll. 2019.
\newblock \href {https://doi.org/10.5210/fm.v24i7.9585} {Managing conflict in
  online debate communities}.
\newblock \emph{First Monday}, 24(7).

\bibitem[{Beel et~al.(2022)Beel, Xiang, Soni, and Yang}]{beel2022linguistic}
Jacob Beel, Tong Xiang, Sandeep Soni, and Diyi Yang. 2022.
\newblock \href {https://doi.org/10.1609/icwsm.v16i1.19270} {Linguistic
  characterization of divisive topics online: Case studies on contentiousness
  in abortion, climate change, and gun control}.
\newblock \emph{Proceedings of the International AAAI Conference on Web and
  Social Media}, 16(1):32--42.

\bibitem[{Bouman et~al.(2018)Bouman, Steg, and Kiers}]{bouman2018measuring}
Thijs Bouman, Linda Steg, and Henk A.~L. Kiers. 2018.
\newblock \href {https://doi.org/10.3389/fpsyg.2018.00564} {Measuring values in
  environmental research: A test of an environmental portrait value
  questionnaire}.
\newblock \emph{Frontiers in Psychology}, 9.

\bibitem[{Boyd et~al.(2021)Boyd, Wilson, Pennebaker, Kosinski, Stillwell, and
  Mihalcea}]{boyd2015values}
Ryan Boyd, Steven Wilson, James Pennebaker, Michal Kosinski, David Stillwell,
  and Rada Mihalcea. 2021.
\newblock \href {https://doi.org/10.1609/icwsm.v9i1.14589} {Values in words:
  Using language to evaluate and understand personal values}.
\newblock \emph{Proceedings of the International AAAI Conference on Web and
  Social Media}, 9(1):31--40.

\bibitem[{Cabitza et~al.(2023)Cabitza, Campagner, and
  Basile}]{cabitza2023toward}
Federico Cabitza, Andrea Campagner, and Valerio Basile. 2023.
\newblock \href {https://doi.org/10.1609/aaai.v37i6.25840} {Toward a
  perspectivist turn in ground truthing for predictive computing}.
\newblock \emph{Proceedings of the AAAI Conference on Artificial Intelligence},
  37(6):6860--6868.

\bibitem[{Chen et~al.(2014)Chen, Hsieh, Mahmud, and
  Nichols}]{chen2014understanding}
Jilin Chen, Gary Hsieh, Jalal~U. Mahmud, and Jeffrey Nichols. 2014.
\newblock \href {https://doi.org/10.1145/2531602.2531608} {Understanding
  individuals' personal values from social media word use}.
\newblock In \emph{Proceedings of the 17th ACM Conference on Computer Supported
  Cooperative Work \& Social Computing}, CSCW '14, page 405–414, New York,
  NY, USA. Association for Computing Machinery.

\bibitem[{Chen et~al.(2019)Chen, Khashabi, Yin, Callison-Burch, and
  Roth}]{chen2019seeing}
Sihao Chen, Daniel Khashabi, Wenpeng Yin, Chris Callison-Burch, and Dan Roth.
  2019.
\newblock \href {https://www.aclweb.org/anthology/N19-1053.pdf} {Seeing things
  from a different angle: Discovering diverse perspectives about claims}.
\newblock In \emph{Proceedings of the 2019 Conference of the North American
  Chapter of the Association for Computational Linguistics: Human Language
  Technologies, Volume 1 (Long and Short Papers)}, pages 542--557.

\bibitem[{Chew et~al.(2021)Chew, Kery, Baum, Bukowski, Kim, and
  Navarro}]{chew2021predicting}
Robert Chew, Caroline Kery, Laura Baum, Thomas Bukowski, Annice Kim, and Mario
  Navarro. 2021.
\newblock \href {https://doi.org/10.2196/25807} {Predicting age groups of
  reddit users based on posting behavior and metadata: Classification model
  development and validation}.
\newblock \emph{JMIR Public Health Surveill}, 7(3):e25807.

\bibitem[{Croux and Dehon(2010)}]{croux2010influence}
Christophe Croux and Catherine Dehon. 2010.
\newblock \href {https://doi.org/10.1007/s10260-010-0142-z} {Influence
  functions of the spearman and kendall correlation measures}.
\newblock \emph{Statistical Methods {\&} Applications}, 19(4):497--515.

\bibitem[{De~Kock and Vlachos(2021)}]{de2021beg}
Christine De~Kock and Andreas Vlachos. 2021.
\newblock \href {https://aclanthology.org/2021.eacl-main.173/} {I beg to
  differ: A study of constructive disagreement in online conversations}.
\newblock In \emph{Proceedings of the 16th Conference of the European Chapter
  of the Association for Computational Linguistics: Main Volume}, pages
  2017--2027.

\bibitem[{Devlin et~al.(2019)Devlin, Chang, Lee, and
  Toutanova}]{devlin2019bert}
Jacob Devlin, Ming-Wei Chang, Kenton Lee, and Kristina Toutanova. 2019.
\newblock \href {https://doi.org/10.18653/v1/N19-1423} {{BERT}: Pre-training of
  deep bidirectional transformers for language understanding}.
\newblock In \emph{Proceedings of the 2019 Conference of the North {A}merican
  Chapter of the Association for Computational Linguistics: Human Language
  Technologies, Volume 1 (Long and Short Papers)}, pages 4171--4186,
  Minneapolis, Minnesota. Association for Computational Linguistics.

\bibitem[{Draws et~al.(2022)Draws, Inel, Tintarev, Baden, and
  Timmermans}]{draws2022comprehensive}
Tim Draws, Oana Inel, Nava Tintarev, Christian Baden, and Benjamin Timmermans.
  2022.
\newblock \href {https://dl.acm.org/doi/abs/10.1145/3498366.3505812}
  {Comprehensive viewpoint representations for a deeper understanding of user
  interactions with debated topics}.
\newblock In \emph{ACM SIGIR Conference on Human Information Interaction and
  Retrieval}, pages 135--145.

\bibitem[{Esau and Friess(2022)}]{esau2022creates}
K~Esau and Dennis Friess. 2022.
\newblock \href {https://doi.org/10.16997/jdd.1021} {What creates listening
  online? exploring reciprocity in online political discussions with relational
  content analysis}.
\newblock \emph{Journal of Deliberative Democracy}, 18(1):1--16.

\bibitem[{Esterling et~al.(2015)Esterling, Fung, and Lee}]{esterling2015much}
Kevin~M Esterling, Archon Fung, and Taeku Lee. 2015.
\newblock \href
  {https://www.tandfonline.com/doi/pdf/10.1080/10584609.2014.969466} {How much
  disagreement is good for democratic deliberation?}
\newblock \emph{Political Communication}, 32(4):529--551.

\bibitem[{Fang et~al.(2019)Fang, Nadeem, Mohtarami, and Glass}]{fang2019neural}
Wei Fang, Moin Nadeem, Mitra Mohtarami, and James Glass. 2019.
\newblock \href {https://doi.org/10.18653/v1/D19-6603} {Neural multi-task
  learning for stance prediction}.
\newblock In \emph{Proceedings of the Second Workshop on Fact Extraction and
  VERification (FEVER)}, pages 13--19, Hong Kong, China. Association for
  Computational Linguistics.

\bibitem[{Fischer et~al.(2022)Fischer, Reedy, Piercy, and
  Thapaliya}]{fischer2022typology}
Ken Fischer, Justin Reedy, Cameron Piercy, and Rashmi Thapaliya. 2022.
\newblock \href {https://delibdemjournal.org/article/id/951/download/pdf/} {A
  typology of reasoning in deliberative processes: A study of the 2010 oregon
  citizens’ initiative review}.
\newblock \emph{Journal of Deliberative Democracy}, 18(2).

\bibitem[{Friedman et~al.(2013)Friedman, Kahn, Borning, and
  Huldtgren}]{friedman2013value}
Batya Friedman, Peter~H. Kahn, Alan Borning, and Alina Huldtgren. 2013.
\newblock \href {https://doi.org/10.1007/978-94-007-7844-3_4} {\emph{Value
  Sensitive Design and Information Systems}}, pages 55--95. Springer
  Netherlands, Dordrecht.

\bibitem[{Ghosh et~al.(2021)Ghosh, Shrivastava, and
  Muresan}]{ghosh2021laughing}
Debanjan Ghosh, Ritvik Shrivastava, and Smaranda Muresan. 2021.
\newblock \href {https://doi.org/10.18653/v1/2021.eacl-main.171} {{``}laughing
  at you or with you{''}: The role of sarcasm in shaping the disagreement
  space}.
\newblock In \emph{Proceedings of the 16th Conference of the European Chapter
  of the Association for Computational Linguistics: Main Volume}, pages
  1998--2010, Online. Association for Computational Linguistics.

\bibitem[{Graham et~al.(2013)Graham, Haidt, Koleva, Motyl, Iyer, Wojcik, and
  Ditto}]{graham2013moral}
Jesse Graham, Jonathan Haidt, Sena Koleva, Matt Motyl, Ravi Iyer, Sean~P.
  Wojcik, and Peter~H. Ditto. 2013.
\newblock \href
  {https://doi.org/https://doi.org/10.1016/B978-0-12-407236-7.00002-4} {Moral
  foundations theory: The pragmatic validity of moral pluralism}.
\newblock In Patricia Devine and Ashby Plant, editors, \emph{Advances in
  Experimental Social Psychology}, volume~47, pages 55--130. Academic Press.

\bibitem[{Graham et~al.(2009)Graham, Haidt, and Nosek}]{graham2009liberals}
Jesse Graham, Jonathan Haidt, and Brian~A Nosek. 2009.
\newblock \href {https://doi.org/https://doi.org/10.1037/a0015141} {Liberals
  and conservatives rely on different sets of moral foundations.}
\newblock \emph{Journal of personality and social psychology},
  96(5):1029--1046.

\bibitem[{Gu and Budhkar(2021)}]{gu2021package}
Ken Gu and Akshay Budhkar. 2021.
\newblock \href {https://doi.org/10.18653/v1/2021.maiworkshop-1.10} {A package
  for learning on tabular and text data with transformers}.
\newblock In \emph{Proceedings of the Third Workshop on Multimodal Artificial
  Intelligence}, pages 69--73, Mexico City, Mexico. Association for
  Computational Linguistics.

\bibitem[{Hadfi and Ito(2022)}]{hadfi2022augmented}
Rafik Hadfi and Takayuki Ito. 2022.
\newblock \href {https://ifaamas.org/Proceedings/aamas2022/pdfs/p1794.pdf}
  {Augmented democratic deliberation: Can conversational agents boost
  deliberation in social media?}
\newblock In \emph{Proceedings of the 21st International Conference on
  Autonomous Agents and Multiagent Systems}, pages 1794--1798.

\bibitem[{Hardalov et~al.(2021)Hardalov, Arora, Nakov, and
  Augenstein}]{hardalov2021cross}
Momchil Hardalov, Arnav Arora, Preslav Nakov, and Isabelle Augenstein. 2021.
\newblock \href {https://doi.org/10.18653/v1/2021.emnlp-main.710} {Cross-domain
  label-adaptive stance detection}.
\newblock In \emph{Proceedings of the 2021 Conference on Empirical Methods in
  Natural Language Processing}, pages 9011--9028, Online and Punta Cana,
  Dominican Republic. Association for Computational Linguistics.

\bibitem[{Hoover et~al.(2020)Hoover, Portillo-Wightman, Yeh, Havaldar, Davani,
  Lin, Kennedy, Atari, Kamel, Mendlen, Moreno, Park, Chang, Chin, Leong, Leung,
  Mirinjian, and Dehghani}]{hoover2020moral}
Joe Hoover, Gwenyth Portillo-Wightman, Leigh Yeh, Shreya Havaldar,
  Aida~Mostafazadeh Davani, Ying Lin, Brendan Kennedy, Mohammad Atari, Zahra
  Kamel, Madelyn Mendlen, Gabriela Moreno, Christina Park, Tingyee~E. Chang,
  Jenna Chin, Christian Leong, Jun~Yen Leung, Arineh Mirinjian, and Morteza
  Dehghani. 2020.
\newblock \href {https://doi.org/10.1177/1948550619876629} {Moral foundations
  twitter corpus: A collection of 35k tweets annotated for moral sentiment}.
\newblock \emph{Social Psychological and Personality Science},
  11(8):1057--1071.

\bibitem[{Jehn(1994)}]{jehn1994enhancing}
Karen~A. Jehn. 1994.
\newblock \href {https://doi.org/10.1108/eb022744} {Enhancing effectiveness: An
  investigation of advantages and disadvantages of value‐based intragroup
  conflict}.
\newblock \emph{International Journal of Conflict Management}, 5(3):223--238.

\bibitem[{Jhaver et~al.(2019)Jhaver, Bruckman, and Gilbert}]{jhaver2019does}
Shagun Jhaver, Amy Bruckman, and Eric Gilbert. 2019.
\newblock \href {https://doi.org/10.1145/3359252} {Does transparency in
  moderation really matter? user behavior after content removal explanations on
  reddit}.
\newblock \emph{Proc. ACM Hum.-Comput. Interact.}, 3(CSCW).

\bibitem[{Joulin et~al.(2017)Joulin, Grave, Bojanowski, and
  Mikolov}]{joulin2017bag}
Armand Joulin, Edouard Grave, Piotr Bojanowski, and Tomas Mikolov. 2017.
\newblock \href {https://aclanthology.org/E17-2068} {Bag of tricks for
  efficient text classification}.
\newblock In \emph{Proceedings of the 15th Conference of the {E}uropean Chapter
  of the Association for Computational Linguistics: Volume 2, Short Papers},
  pages 427--431, Valencia, Spain. Association for Computational Linguistics.

\bibitem[{Kass and Raftery(1995)}]{kass1995bayes}
Robert~E. Kass and Adrian~E. Raftery. 1995.
\newblock \href {https://doi.org/10.1080/01621459.1995.10476572} {Bayes
  factors}.
\newblock \emph{Journal of the American Statistical Association},
  90(430):773--795.

\bibitem[{Kennedy et~al.(2021)Kennedy, Atari, Davani, Hoover, Omrani, Graham,
  and Dehghani}]{kennedy2021moral}
Brendan Kennedy, Mohammad Atari, Aida~Mostafazadeh Davani, Joe Hoover, Ali
  Omrani, Jesse Graham, and Morteza Dehghani. 2021.
\newblock \href
  {https://www.sciencedirect.com/science/article/pii/S0010027721001153?casa_token=BMUhrR9OfO0AAAAA:cR2i4vGGTEb3lPF8bvQv-H9UYNdcngYOwBM1etcugwbqC5cW7qUhGO9YQmoh0UOtYDA1XD8NGWc}
  {Moral concerns are differentially observable in language}.
\newblock \emph{Cognition}, 212:104696.

\bibitem[{Kiesel et~al.(2022)Kiesel, Alshomary, Handke, Cai, Wachsmuth, and
  Stein}]{kiesel2022identifying}
Johannes Kiesel, Milad Alshomary, Nicolas Handke, Xiaoni Cai, Henning
  Wachsmuth, and Benno Stein. 2022.
\newblock \href {https://aclanthology.org/2022.acl-long.306/} {Identifying the
  human values behind arguments}.
\newblock In \emph{Proceedings of the 60th Annual Meeting of the Association
  for Computational Linguistics (Volume 1: Long Papers)}, pages 4459--4471.

\bibitem[{Kiesel et~al.(2023)Kiesel, Alshomary, Mirzakhmedova, Heinrich,
  Handke, Wachsmuth, and Stein}]{kiesel2023semeval}
Johannes Kiesel, Milad Alshomary, Nailia Mirzakhmedova, Maximilian Heinrich,
  Nicolas Handke, Henning Wachsmuth, and Benno Stein. 2023.
\newblock \href {https://doi.org/10.18653/v1/2023.semeval-1.313}
  {{S}em{E}val-2023 task 4: {V}alue{E}val: Identification of human values
  behind arguments}.
\newblock In \emph{Proceedings of the 17th International Workshop on Semantic
  Evaluation (SemEval-2023)}, pages 2287--2303, Toronto, Canada. Association
  for Computational Linguistics.

\bibitem[{Klein(2012)}]{klein2012enabling}
Mark Klein. 2012.
\newblock \href {https://doi.org/10.2139/ssrn.1837707} {Enabling large-scale
  deliberation using attention-mediation metrics}.
\newblock \emph{Computer Supported Cooperative Work (CSCW)}, 21(4-5):449--473.

\bibitem[{Kobbe et~al.(2020)Kobbe, Rehbein, Hulpu{\textcommabelow{s}}, and
  Stuckenschmidt}]{kobbe2020exploring}
Jonathan Kobbe, Ines Rehbein, Ioana Hulpu{\textcommabelow{s}}, and Heiner
  Stuckenschmidt. 2020.
\newblock \href {https://aclanthology.org/2020.argmining-1.4} {Exploring
  morality in argumentation}.
\newblock In \emph{Proceedings of the 7th Workshop on Argument Mining}, pages
  30--40, Online. Association for Computational Linguistics.

\bibitem[{Kruschke(2018)}]{kruschke2018rejecting}
John~K. Kruschke. 2018.
\newblock \href {https://doi.org/10.1177/2515245918771304} {Rejecting or
  accepting parameter values in bayesian estimation}.
\newblock \emph{Advances in Methods and Practices in Psychological Science},
  1(2):270--280.

\bibitem[{Lin and Kim(2023)}]{lin2022learning}
Han Lin and Yonghwan Kim. 2023.
\newblock \href {https://doi.org/https://doi.org/10.1016/j.chb.2022.107558}
  {Learning from disagreement on social media: The mediating role of
  like-minded and cross-cutting discussion and the moderating role of
  fact-checking}.
\newblock \emph{Computers in Human Behavior}, 139:107558.

\bibitem[{Liscio et~al.(2023)Liscio, Araque, Gatti, Constantinescu, Jonker,
  Kalimeri, and Murukannaiah}]{liscio2023does}
Enrico Liscio, Oscar Araque, Lorenzo Gatti, Ionut Constantinescu, Catholijn
  Jonker, Kyriaki Kalimeri, and Pradeep~Kumar Murukannaiah. 2023.
\newblock \href {https://doi.org/10.18653/v1/2023.acl-long.789} {What does a
  text classifier learn about morality? an explainable method for cross-domain
  comparison of moral rhetoric}.
\newblock In \emph{Proceedings of the 61st Annual Meeting of the Association
  for Computational Linguistics (Volume 1: Long Papers)}, pages 14113--14132,
  Toronto, Canada. Association for Computational Linguistics.

\bibitem[{Liscio et~al.(2022{\natexlab{a}})Liscio, Dondera, Geadau, Jonker, and
  Murukannaiah}]{liscio2022cross}
Enrico Liscio, Alin~E Dondera, Andrei Geadau, Catholijn~M Jonker, and Pradeep~K
  Murukannaiah. 2022{\natexlab{a}}.
\newblock \href
  {https://enricoliscio.github.io/publications/Liscio_cross_domain_values_NAACL.pdf}
  {Cross-domain classification of moral values}.
\newblock In \emph{Findings of the 2022 Conference of the North American
  Chapter of the Association for Computational Linguistics. NAACL}, volume~22,
  pages 1--13.

\bibitem[{Liscio et~al.(2021)Liscio, van~der Meer, Siebert, Jonker, Mouter, and
  Murukannaiah}]{Liscio-2021-AAMAS-Axies}
Enrico Liscio, Michiel van~der Meer, Luciano~C. Siebert, Catholijn~M. Jonker,
  Niek Mouter, and Pradeep~K. Murukannaiah. 2021.
\newblock \href
  {https://ii.tudelft.nl/~pradeep/doc/Liscio-2021-AAMAS-Axies.pdf} {Axies:
  Identifying and evaluating context-specific values}.
\newblock In \emph{Proceedings of the 20th Conference on Autonomous Agents and
  MultiAgent Systems}, AAMAS '21, pages 799--808, London.

\bibitem[{Liscio et~al.(2022{\natexlab{b}})Liscio, van~der Meer, Siebert,
  Jonker, and Murukannaiah}]{liscio2022values}
Enrico Liscio, Michiel van~der Meer, Luciano~C. Siebert, Catholijn~M. Jonker,
  and Pradeep~K. Murukannaiah. 2022{\natexlab{b}}.
\newblock \href {https://doi.org/10.1007/s10458-022-09550-0} {What values
  should an agent align with? {A}n empirical comparison of general and
  context-specific values}.
\newblock \emph{Autonomous Agents and Multi-Agent Systems}, 36(1):23.

\bibitem[{Liu et~al.(2021)Liu, Lin, Tan, and Wang}]{liu2021enhancing}
Rui Liu, Zheng Lin, Yutong Tan, and Weiping Wang. 2021.
\newblock \href {https://doi.org/10.18653/v1/2021.findings-acl.278} {Enhancing
  zero-shot and few-shot stance detection with commonsense knowledge graph}.
\newblock In \emph{Findings of the Association for Computational Linguistics:
  ACL-IJCNLP 2021}, pages 3152--3157, Online. Association for Computational
  Linguistics.

\bibitem[{Liu and Lapata(2019)}]{liu2019text}
Yang Liu and Mirella Lapata. 2019.
\newblock \href {https://doi.org/10.18653/v1/d19-1387} {Text summarization with
  pretrained encoders}.
\newblock \emph{arXiv preprint arXiv:1908.08345}.

\bibitem[{Lourie et~al.(2021)Lourie, Le~Bras, and Choi}]{lourie2021scruples}
Nicholas Lourie, Ronan Le~Bras, and Yejin Choi. 2021.
\newblock \href {https://doi.org/10.1609/aaai.v35i15.17589} {Scruples: A corpus
  of community ethical judgments on 32,000 real-life anecdotes}.
\newblock \emph{Proceedings of the AAAI Conference on Artificial Intelligence},
  35(15):13470--13479.

\bibitem[{Luo et~al.(2023)Luo, Liu, Li, and Zhang}]{luo2023improving}
Yun Luo, Zihan Liu, Stan~Z. Li, and Yue Zhang. 2023.
\newblock \href {https://doi.org/10.1145/3543507.3583314} {Improving
  (dis)agreement detection with inductive social relation information from
  comment-reply interactions}.
\newblock In \emph{Proceedings of the ACM Web Conference 2023}, WWW '23, page
  1584–1593, New York, NY, USA. Association for Computing Machinery.

\bibitem[{Mehrabi et~al.(2021)Mehrabi, Morstatter, Saxena, Lerman, and
  Galstyan}]{mehrabi2021survey}
Ninareh Mehrabi, Fred Morstatter, Nripsuta Saxena, Kristina Lerman, and Aram
  Galstyan. 2021.
\newblock \href {https://doi.org/10.1145/3457607} {A survey on bias and
  fairness in machine learning}.
\newblock \emph{ACM Comput. Surv.}, 54(6).

\bibitem[{Moerbeek(2021)}]{moerbeek2021bayesian}
Mirjam Moerbeek. 2021.
\newblock \href {https://doi.org/10.1186/s12874-021-01334-6} {Bayesian
  updating: increasing sample size during the course of a study}.
\newblock \emph{BMC Medical Research Methodology}, 21(1):137.

\bibitem[{Montani and Honnibal(2022)}]{prodigy_montani_honnibal}
Ines Montani and Matthew Honnibal. 2022.
\newblock \href {https://prodi.gy/} {Prodigy: A modern and scriptable
  annotation tool for creating training data for machine learning models}.

\bibitem[{Niculae and
  Danescu-Niculescu-Mizil(2016)}]{niculae2016conversational}
Vlad Niculae and Cristian Danescu-Niculescu-Mizil. 2016.
\newblock \href {https://aclanthology.org/N16-1070.pdf} {Conversational markers
  of constructive discussions}.
\newblock In \emph{Proceedings of the 2016 Conference of the North American
  Chapter of the Association for Computational Linguistics: Human Language
  Technologies}, pages 568--578.

\bibitem[{Obie et~al.(2021)Obie, Hussain, Xia, Grundy, Li, Turhan, Whittle, and
  Shahin}]{obie2021first}
Humphrey~O. Obie, Waqar Hussain, Xin Xia, John Grundy, Li~Li, Burak Turhan, Jon
  Whittle, and Mojtaba Shahin. 2021.
\newblock \href {https://doi.org/10.1109/ICSE-SEIS52602.2021.00012} {A first
  look at human values-violation in app reviews}.
\newblock In \emph{2021 IEEE/ACM 43rd International Conference on Software
  Engineering: Software Engineering in Society (ICSE-SEIS)}, pages 29--38.

\bibitem[{Pedregosa et~al.(2011)Pedregosa, Varoquaux, Gramfort, Michel,
  Thirion, Grisel, Blondel, Prettenhofer, Weiss, Dubourg, Vanderplas, Passos,
  Cournapeau, Brucher, Perrot, and {{\'E}}douard Duchesnay}]{scikit-learn}
Fabian Pedregosa, Ga{{\"e}}l Varoquaux, Alexandre Gramfort, Vincent Michel,
  Bertrand Thirion, Olivier Grisel, Mathieu Blondel, Peter Prettenhofer, Ron
  Weiss, Vincent Dubourg, Jake Vanderplas, Alexandre Passos, David Cournapeau,
  Matthieu Brucher, Matthieu Perrot, and {{\'E}}douard Duchesnay. 2011.
\newblock \href {http://jmlr.org/papers/v12/pedregosa11a.html} {Scikit-learn:
  Machine learning in python}.
\newblock \emph{Journal of Machine Learning Research}, 12(85):2825--2830.

\bibitem[{Polletta and Gardner(2018)}]{polletta2018forms}
Francesca Polletta and Beth Gardner. 2018.
\newblock \href {https://doi.org/10.1093/oxfordhb/9780198747369.013.45} {{The
  Forms of Deliberative Communication}}.
\newblock In \emph{{The Oxford Handbook of Deliberative Democracy}}. Oxford
  University Press.

\bibitem[{Ponizovskiy et~al.(2020)Ponizovskiy, Ardag, Grigoryan, Boyd,
  Dobewall, and Holtz}]{ponizovskiy2020development}
Vladimir Ponizovskiy, Murat Ardag, Lusine Grigoryan, Ryan Boyd, Henrik
  Dobewall, and Peter Holtz. 2020.
\newblock \href {https://journals.sagepub.com/doi/pdf/10.1002/per.2294}
  {Development and validation of the personal values dictionary: A
  theory--driven tool for investigating references to basic human values in
  text}.
\newblock \emph{European Journal of Personality}, 34(5):885--902.

\bibitem[{Pougu{\'e}-Biyong et~al.(2021)Pougu{\'e}-Biyong, Semenova, Matton,
  Han, Kim, Lambiotte, and Farmer}]{pougue2021debagreement}
John Pougu{\'e}-Biyong, Valentina Semenova, Alexandre Matton, Rachel Han, Aerin
  Kim, Renaud Lambiotte, and Doyne Farmer. 2021.
\newblock \href {https://openreview.net/pdf?id=udVUN__gFO} {Debagreement: A
  comment-reply dataset for (dis) agreement detection in online debates}.
\newblock In \emph{Thirty-fifth Conference on Neural Information Processing
  Systems Datasets and Benchmarks Track (Round 2)}.

\bibitem[{Proferes et~al.(2021)Proferes, Jones, Gilbert, Fiesler, and
  Zimmer}]{proferes2021studying}
Nicholas Proferes, Naiyan Jones, Sarah Gilbert, Casey Fiesler, and Michael
  Zimmer. 2021.
\newblock \href {https://doi.org/10.1177/20563051211019004} {Studying {R}eddit:
  A systematic overview of disciplines, approaches, methods, and ethics}.
\newblock \emph{Social Media + Society}, 7(2):20563051211019004.

\bibitem[{Qiu et~al.(2021)Qiu, Zhao, Li, Lu, Peng, Gao, and
  Zhu}]{qiu2021valuenet}
Liang Qiu, Yizhou Zhao, Jinchao Li, Pan Lu, Baolin Peng, Jianfeng Gao, and
  Song-Chun Zhu. 2021.
\newblock \href {https://arxiv.org/pdf/2112.06346.pdf} {Valuenet: A new dataset
  for human value driven dialogue system}.
\newblock \emph{arXiv preprint arXiv:2112.06346}.

\bibitem[{Rouder et~al.(2009)Rouder, Speckman, Sun, Morey, and
  Iverson}]{rouder2009bayesian}
Jeffrey~N. Rouder, Paul~L. Speckman, Dongchu Sun, Richard~D. Morey, and
  Geoffrey Iverson. 2009.
\newblock \href {https://doi.org/10.3758/PBR.16.2.225} {Bayesian t tests for
  accepting and rejecting the null hypothesis}.
\newblock \emph{Psychonomic Bulletin {\&} Review}, 16(2):225--237.

\bibitem[{Saveski et~al.(2022)Saveski, Gillani, Yuan, Vijayaraghavan, and
  Roy}]{saveski2022perspective}
Martin Saveski, Nabeel Gillani, Ann Yuan, Prashanth Vijayaraghavan, and Deb
  Roy. 2022.
\newblock \href {https://ojs.aaai.org/index.php/ICWSM/article/view/19343/19115}
  {Perspective-taking to reduce affective polarization on social media}.
\newblock In \emph{Proceedings of the International AAAI Conference on Web and
  Social Media}, volume~16, pages 885--895.

\bibitem[{Schwartz(2021)}]{schwartz2021repository}
Shalom Schwartz. 2021.
\newblock \href {https://doi.org/10.9707/2307-0919.1173} {A repository of
  schwartz value scales with instructions and an introduction}.
\newblock \emph{Online Readings in Psychology and Culture}, 2.

\bibitem[{Schwartz(2012)}]{schwartz2012overview}
Shalom~H Schwartz. 2012.
\newblock \href {https://doi.org/10.9707/2307-0919.1116} {An overview of the
  schwartz theory of basic values}.
\newblock \emph{Online readings in Psychology and Culture}, 2(1):2307--0919.

\bibitem[{Schwartz et~al.(2012)Schwartz, Cieciuch, Vecchione, Davidov, Fischer,
  Beierlein, Ramos, Verkasalo, L{\"o}nnqvist, Demirutku
  et~al.}]{schwartz2012refining}
Shalom~H Schwartz, Jan Cieciuch, Michele Vecchione, Eldad Davidov, Ronald
  Fischer, Constanze Beierlein, Alice Ramos, Markku Verkasalo, Jan-Erik
  L{\"o}nnqvist, Kursad Demirutku, et~al. 2012.
\newblock \href {https://doi.org/https://doi.org/10.1037/a0029393} {Refining
  the theory of basic individual values.}
\newblock \emph{Journal of personality and social psychology}, 103(4):663.

\bibitem[{Shortall et~al.(2022)Shortall, Itten, van~der Meer, Murukannaiah, and
  Jonker}]{shortall2022reason}
Ruth Shortall, Anatol Itten, Michiel van~der Meer, Pradeep Murukannaiah, and
  Catholijn Jonker. 2022.
\newblock \href
  {https://www.frontiersin.org/articles/10.3389/fpos.2022.946589/full} {Reason
  against the machine? {F}uture directions for mass online deliberation}.
\newblock \emph{Frontiers in Political Science}.

\bibitem[{Siebert et~al.(2022)Siebert, Liscio, Murukannaiah, Kaptein, Spruit,
  van~den Hoven, and Jonker}]{siebert2022estimating}
Luciano~C Siebert, Enrico Liscio, Pradeep~K Murukannaiah, Lionel Kaptein,
  Shannon Spruit, Jeroen van~den Hoven, and Catholijn Jonker. 2022.
\newblock \href
  {https://www.hhai-conference.org/wp-content/uploads/2022/06/hhai-2022_paper_31.pdf}
  {Estimating value preferences in a hybrid participatory system}.
\newblock In \emph{Proceedings of the first International Conference on Hybrid
  Human-Artificial Intelligence (HHAI 2022)}, pages 1--14, Amsterdam, the
  Netherlands. IOS Press.

\bibitem[{Silva et~al.(2021)Silva, Lo, and Lim}]{silva2021predicting}
Amila Silva, Pei-Chi Lo, and Ee~Peng Lim. 2021.
\newblock \href
  {https://ojs.aaai.org/index.php/ICWSM/article/download/18094/17897/21589} {On
  predicting personal values of social media users using community-specific
  language features and personal value correlation}.
\newblock In \emph{Proceedings of the International AAAI Conference on Web and
  Social Media}, volume~15, pages 680--690.

\bibitem[{S{\o}gaard et~al.(2021)S{\o}gaard, Ebert, Bastings, and
  Filippova}]{sogaard2021we}
Anders S{\o}gaard, Sebastian Ebert, Jasmijn Bastings, and Katja Filippova.
  2021.
\newblock \href {https://doi.org/10.18653/v1/2021.eacl-main.156} {We need to
  talk about random splits}.
\newblock In \emph{Proceedings of the 16th Conference of the European Chapter
  of the Association for Computational Linguistics: Main Volume}, pages
  1823--1832, Online. Association for Computational Linguistics.

\bibitem[{Stromer-Galley et~al.(2020)Stromer-Galley, Bryant, and
  Bimber}]{stromer2020context}
Jennifer Stromer-Galley, Lauren Bryant, and Bruce Bimber. 2020.
\newblock \href {https://doi.org/https://doi.org/10.16997/jdd.218} {Context and
  medium matter: Expressing disagreements online and face-to-face in political
  deliberations}.
\newblock \emph{Journal of Deliberative Democracy}, 11(1).

\bibitem[{Stromer-Galley and Muhlberger(2009)}]{stromer2009agreement}
Jennifer Stromer-Galley and Peter Muhlberger. 2009.
\newblock \href {https://doi.org/10.1080/10584600902850775} {Agreement and
  disagreement in group deliberation: Effects on deliberation satisfaction,
  future engagement, and decision legitimacy}.
\newblock \emph{Political Communication}, 26(2):173--192.

\bibitem[{Trager et~al.(2022)Trager, Ziabari, Davani, Golazazian,
  Karimi-Malekabadi, Omrani, Li, Kennedy, Reimer, Reyes
  et~al.}]{trager2022moral}
Jackson Trager, Alireza~S Ziabari, Aida~Mostafazadeh Davani, Preni Golazazian,
  Farzan Karimi-Malekabadi, Ali Omrani, Zhihe Li, Brendan Kennedy, Nils~Karl
  Reimer, Melissa Reyes, et~al. 2022.
\newblock \href {https://arxiv.org/pdf/2208.05545.pdf} {The moral foundations
  reddit corpus}.
\newblock \emph{arXiv preprint arXiv:2208.05545}.

\bibitem[{Vallat(2018)}]{vallat2018pingouin}
Raphael Vallat. 2018.
\newblock \href {https://doi.org/10.21105/joss.01026} {Pingouin: statistics in
  python}.
\newblock \emph{Journal of Open Source Software}, 3(31):1026.

\bibitem[{van~der Meer et~al.(2022{\natexlab{a}})van~der Meer, Liscio, Jonker,
  Plaat, Vossen, and Murukannaiah}]{vandermeer2022hyena}
Michiel van~der Meer, Enrico Liscio, Catholijn~M. Jonker, Aske Plaat, Piek
  Vossen, and Pradeep~K. Murukannaiah. 2022{\natexlab{a}}.
\newblock \href {https://liacs.leidenuniv.nl/~meermtvander/publications/hyena/}
  {{HyEnA: A Hybrid Method for Extracting Arguments from Opinions}}.
\newblock In \emph{Proceedings of the first International Conference on Hybrid
  Human-Artificial Intelligence (HHAI 2022)}, pages 1--15, Amsterdam, the
  Netherlands. IOS Press.

\bibitem[{van~der Meer et~al.(2022{\natexlab{b}})van~der Meer, Reuver, Khurana,
  Krause, and Baez~Santamaria}]{vandermeer2022will}
Michiel van~der Meer, Myrthe Reuver, Urja Khurana, Lea Krause, and Selene
  Baez~Santamaria. 2022{\natexlab{b}}.
\newblock \href {https://aclanthology.org/2022.argmining-1.8} {Will it blend?
  {M}ixing training paradigms {\&} prompting for argument quality prediction}.
\newblock In \emph{Proceedings of the 9th Workshop on Argument Mining}, pages
  95--103, Online and in Gyeongju, Republic of Korea. International Conference
  on Computational Linguistics.

\bibitem[{van~der Meer et~al.(2023)van~der Meer, Vossen, Jonker, and
  Murukannaiah}]{vandermeer2023do_data}
Michiel van~der Meer, Piek Vossen, Catholijn~M. Jonker, and Pradeep~K.
  Murukannaiah. 2023.
\newblock \href {https://doi.org/10.17605/OSF.IO/42DNS} {Do differences in
  values influence disagreements in online discussions? {S}upplementary
  material}.

\bibitem[{van Ravenzwaaij and Wagenmakers(2022)}]{van2022advantages}
Don van Ravenzwaaij and Eric-Jan Wagenmakers. 2022.
\newblock \href {https://doi.org/10.1037/met0000415} {Advantages masquerading
  as {\textquotedblleft}issues{\textquotedblright} in bayesian hypothesis
  testing: A commentary on tendeiro and kiers (2019).}
\newblock \emph{Psychological Methods}, 27(3):451--465.

\bibitem[{Vecchi et~al.(2021)Vecchi, Falk, Jundi, and
  Lapesa}]{vecchi2021towards}
Eva~Maria Vecchi, Neele Falk, Iman Jundi, and Gabriella Lapesa. 2021.
\newblock \href {https://aclanthology.org/2021.acl-long.107.pdf} {Towards
  argument mining for social good: A survey}.
\newblock In \emph{Proceedings of the 59th Annual Meeting of the Association
  for Computational Linguistics and the 11th International Joint Conference on
  Natural Language Processing (Volume 1: Long Papers)}, pages 1338--1352.

\bibitem[{Weinzierl and Harabagiu(2022)}]{weinzierl2022hesitancy}
Maxwell~A Weinzierl and Sanda~M Harabagiu. 2022.
\newblock \href {https://ojs.aaai.org/index.php/ICWSM/article/view/19360/19132}
  {From hesitancy framings to vaccine hesitancy profiles: A journey of stance,
  ontological commitments and moral foundations}.
\newblock In \emph{Proceedings of the International AAAI Conference on Web and
  Social Media}, volume~16, pages 1087--1097.

\bibitem[{Wojatzki et~al.(2018)Wojatzki, Zesch, Mohammad, and
  Kiritchenko}]{wojatzki2018agree}
Michael Wojatzki, Torsten Zesch, Saif Mohammad, and Svetlana Kiritchenko. 2018.
\newblock \href {https://aclanthology.org/S18-2026.pdf} {Agree or disagree:
  Predicting judgments on nuanced assertions}.
\newblock In \emph{Proceedings of the Seventh Joint Conference on Lexical and
  Computational Semantics}, pages 214--224.

\bibitem[{Wolf et~al.(2020)Wolf, Debut, Sanh, Chaumond, Delangue, Moi, Cistac,
  Rault, Louf, Funtowicz, Davison, Shleifer, von Platen, Ma, Jernite, Plu, Xu,
  Scao, Gugger, Drame, Lhoest, and Rush}]{wolf2020transformers}
Thomas Wolf, Lysandre Debut, Victor Sanh, Julien Chaumond, Clement Delangue,
  Anthony Moi, Pierric Cistac, Tim Rault, Rémi Louf, Morgan Funtowicz, Joe
  Davison, Sam Shleifer, Patrick von Platen, Clara Ma, Yacine Jernite, Julien
  Plu, Canwen Xu, Teven~Le Scao, Sylvain Gugger, Mariama Drame, Quentin Lhoest,
  and Alexander~M. Rush. 2020.
\newblock \href {https://www.aclweb.org/anthology/2020.emnlp-demos.6}
  {Transformers: State-of-the-art natural language processing}.
\newblock In \emph{Proceedings of the 2020 Conference on Empirical Methods in
  Natural Language Processing: System Demonstrations}, pages 38--45, Online.
  Association for Computational Linguistics.

\end{thebibliography}
